\documentclass{article} 

\usepackage{amsmath,amsfonts,bm}

\def\eqref#1{equation~\ref{#1}}

\def\1{\bm{1}}

\DeclareMathAlphabet{\mathsfit}{\encodingdefault}{\sfdefault}{m}{sl}
\SetMathAlphabet{\mathsfit}{bold}{\encodingdefault}{\sfdefault}{bx}{n}

\usepackage{fullpage}
\usepackage{hyperref}
\hypersetup{colorlinks,linkcolor={cvprblue},citecolor={cvprblue},urlcolor={RoyalBlue}} 
\usepackage{url}
\usepackage{graphicx}
\usepackage{cleveref}
\usepackage{booktabs}       
\usepackage{xcolor}         
\usepackage{float}
\usepackage{colortbl}
\usepackage{enumitem}
\usepackage{multirow}
\usepackage[mathscr]{eucal}
\usepackage[export]{adjustbox}
\usepackage{multicol,lipsum}
\usepackage{arydshln}
\usepackage{pifont}
\usepackage{bbding}
\usepackage[table, dvipsnames, svgnames]{xcolor}
\usepackage{comment}
\usepackage{wrapfig}
\usepackage{subcaption}
\usepackage{makecell}
\usepackage{pifont}
\usepackage[numbers]{natbib}

\usepackage[utf8]{inputenc} 
\usepackage[T1]{fontenc}    
\usepackage{hyperref}
\usepackage{xspace}
\usepackage{url}            
\usepackage{graphicx}
\usepackage{multirow}
\usepackage{colortbl}
\usepackage{caption, subcaption}
\usepackage{booktabs}  
\usepackage{arydshln}  
\usepackage{xcolor}
\usepackage{enumitem}
\usepackage[export]{adjustbox}
\usepackage{multicol,lipsum}
\usepackage{makecell}

\usepackage{amsmath}  
\usepackage{wrapfig}
\usepackage{microtype}
\usepackage{pifont}  

\usepackage{array}
\usepackage{tikz}

\definecolor{cvprblue}{rgb}{0.21,0.49,0.74}
\definecolor{purple}{rgb}{0.56,0.27,0.68}
\definecolor{red}{rgb}{0.95,0.4,0.4}
\definecolor{purered}{rgb}{1,0,0}
\definecolor{blue}{rgb}{0.4,0.4,0.95}
\definecolor{darkblue}{rgb}{0,0,0.8}
\definecolor{lightblue}{rgb}{127,153,240}
\definecolor{grey}{rgb}{0.6,0.6,0.6}
\definecolor{col1}{RGB}{233, 255, 245}

\definecolor{col11}{RGB}{255, 228, 228}
\definecolor{col2}{RGB}{148, 187, 232}
\definecolor{col33}{RGB}{206, 239, 255}
\definecolor{col3}{RGB}{225, 239, 220} 
\definecolor{lightgrey}{rgb}{0.85,0.85,0.85}
\definecolor{lightyellow}{RGB}{255,195,78}
\definecolor{lightlightgrey}{rgb}{0.9,0.9,0.9}
\definecolor{verylightBG}{rgb}{0.9,0.99,0.99}
\definecolor{darkgreen}{rgb}{0., 0.85, 0.5}
\definecolor{gtred}{RGB}{204, 0, 0}
\definecolor{predgreen}{RGB}{31, 237, 31}
\definecolor{figGreen}{RGB}{56, 118, 29}

\definecolor{colorNoSelection}{RGB}{227, 155, 0}

\definecolor{gFONE}{RGB}{154, 113, 196}

\definecolor{IDtrn}{RGB}{140, 85, 74}

\definecolor{Rtr}{RGB}{44, 160, 44}

\definecolor{ArithmeticMean}{RGB}{188, 189, 33}

\definecolor{BordaCount}{RGB}{90, 169, 162}

\definecolor{FTWO}{RGB}{231, 110, 141}

\definecolor{FHALF}{RGB}{231, 144, 141}

\let\citeoriginal\cite
\renewcommand{\cite}[1]{{\normalsize\citeoriginal{#1}}}

\newcommand{\revised}[1]{{\leavevmode\color[rgb]{0.,0.,0.}#1}}

\newcommand*{\ourmodel}{SAYD\@\xspace}

\title{Segment Anything In-Context via  Visual Demonstrations}

\author{Yuzhu Wang \quad Manni Duan \quad Hongliang Ye \quad Shu Kong \\
{\small \em code: \url{https://github.com/WangYZ1608/SAYD}}
\vspace{-3mm}
}

\begin{document}

\maketitle

\begin{abstract}
We present the \textit{Segment Anything Model via Visual Demonstrations} (\ourmodel),
a foundation model for in-context segmentation.
Given visual demos comprising prompt images and their segmentation masks,
\ourmodel segments coherent regions (e.g., objects or their parts) in a query image by establishing visual correspondences between the prompts and the query.
\ourmodel generalizes to previously-unseen segmentation tasks, including those requiring tacit visual knowledge or involving concepts that are difficult to articulate unambiguously in language.
We train \ourmodel on 12 public datasets originally curated for panoptic segmentation using our proposed techniques, including a lightweight mask decoder for prompt-agnostic mask generation and a learning mechanism for recognizing novel query classes absent in the prompts.
These innovations enhance both computational efficiency and out-of-domain generalization.
Consequently, \ourmodel significantly outperforms existing methods of in-context segmentation and open-vocabulary segmentation across 24 out-of-domain benchmarks,
spanning natural, biological, ecological, cosmological, medical, pathological, and satellite imagery.

\end{abstract}

\begin{figure}[h]%
\centering
\vspace{-1mm}
\includegraphics[width=\linewidth]{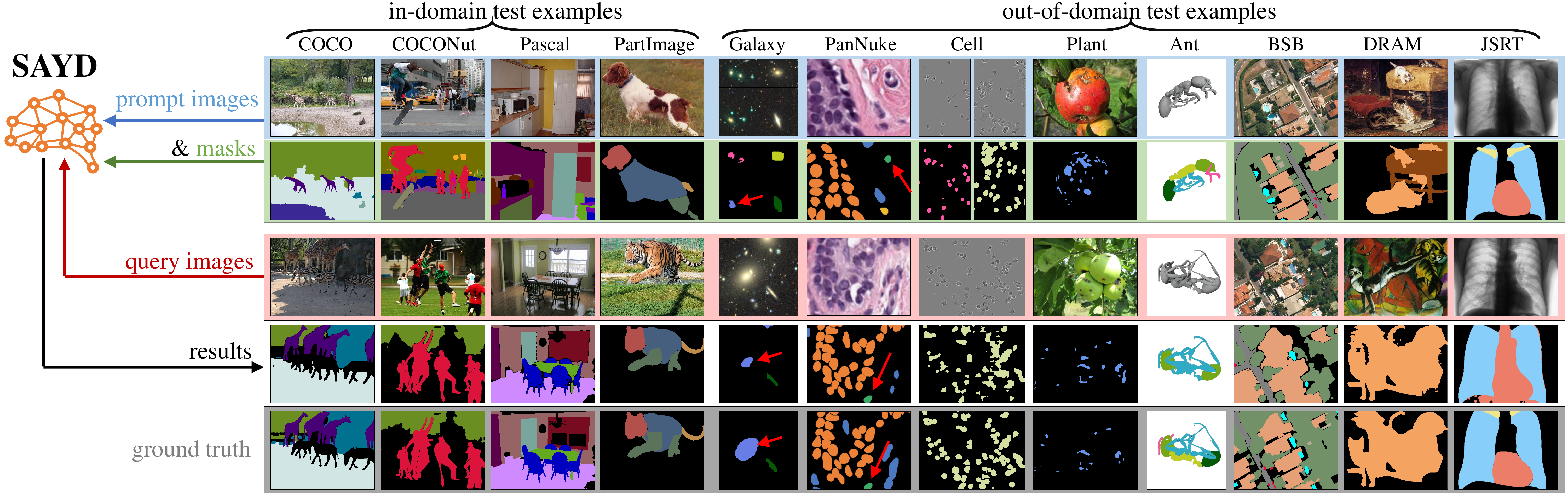}
\vspace{-5mm}
\caption{\small 
Our foundation model \ourmodel solves in-context segmentation.
We train it on public datasets originally curated for panoptic segmentation (ref. the first four columns showing test images from these datasets).
\ourmodel generalizes to out-of-domain benchmarks spanning natural, biological, ecological, cosmological, medical,
pathological, and satellite imagery (ref. examples shown in the last eight columns).
\ourmodel effectively segments the visual targets,
particularly those that are often identified by domain experts through tacit visual knowledge, indicated by the red arrow in the fifth and sixth columns.
}
\label{fig:splashing_figure}
\vspace{-3mm}
\end{figure}

\section{Introduction}
\label{sec:intro}

The ability to segment coherent regions (e.g., objects or their parts) in an image based on visual demonstrations is intrinsic to human learning  \cite{meltzoff1999imitation}.
This capability emerges prior to the acquisition of high-level semantic knowledge, which is typically conveyed through language \cite{lavi2018developmental, johnson2010infants}.
Recent computer vision literature has formalized this ability as In-Context Segmentation (ICS) \cite{bar2022visual, cuttano2026insid3},
aiming to segment a query image conditioned on visual demos comprising prompt images and their corresponding segmentation masks.
Unlike the language-based prompts used by prevailing foundational segmentation models, such as SAM3 \cite{carion2025sam}, 
visual demos can naturally encapsulate concepts rooted in tacit visual knowledge \cite{collins1974tea, heiberg2008tacit, patel1999expertise}.
ICS is therefore particularly valuable for domains constrained by the knowledge acquisition bottleneck \cite{hanney1997adaptation, lenat1985cyc},
where relevant concepts may be difficult to articulate unambiguously.
To solve ICS, we present a foundation model termed \textit{\underline{S}egment \underline{A}n\underline{y}thing Model via Visual \underline{D}emonstrations (SAYD)}, as illustrated in Fig.~\ref{fig:splashing_figure}.

\begin{figure}[t]
    \centering
    \vspace{-2mm}
        \includegraphics[width=0.999\textwidth]{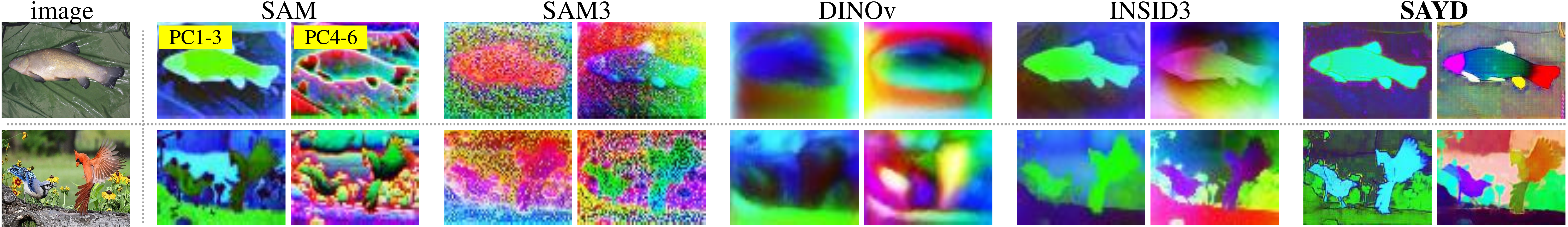}
\vspace{-7mm}
\caption{
\small
{\bf Visualization of feature maps from different models.}
Given an image, we extract the feature maps using the models and display their first and second three principal components as RGB images. 
Notably, INSID3 exploits the frozen DINOv3 as the image encoder while \ourmodel finetunes it for ICS.
\ourmodel exhibits clear hierarchical pixel grouping ability across different levels of granularity, intuitively explaining its learned representations effective for ICS.
}
\vspace{-1mm}
\label{fig:pca}
\end{figure}

Early ICS methods do not directly solve ICS but train proxy models, such as pixel inpainting \cite{bar2022visual} and pixel coloring
\cite{wang2023images, wang2023seggpt}, and subsequently post-process their outputs into segmentation masks.
Most recent approaches exploit frozen pretrained image encoders \cite{balazevic2023towards, liu2023matcher, meng2024segic, liu2024simple, cuttano2026insid3}, focusing instead on matching off-the-shelf features between the prompt and query images toward ICS.
As the encoders are not pretrained specifically for ICS,
these methods are limited in their matching ability
required for effective ICS (Fig.~\ref{fig:pca}).
DINOv~\cite{li2024visual} is a universal visual in-context prompting model trained on 11M 
images,
supporting not only ICS but also other prompts (e.g., strokes, boxes, and points).
However, 
its focus on universal prompts hinders its optimization for broad ICS deployment. 
In sum, existing ICS methods struggle to generalize to diverse applications, particularly in scientific imaging and are thus ill-equipped to serve as foundation models (FMs) for these domains (Fig.~\ref{fig:quick-visual-comparison}).

To address these limitations, we introduce the foundation model \ourmodel. 
First,
we exploit diverse public datasets curated for various panoptic segmentation tasks, which provide class labels and masks for objects and/or parts.
Using them not only enriches training but also supports a controllable reproducible evaluation on out-of-domain (OOD) generalization \cite{shi2024lca} (\S\ref{sec:problem}). 
Second, to enable effective training across heterogeneous datasets,
we propose several simple and novel techniques (\S\ref{sec:method}) regarding architecture design, training data sampling, and loss formulation.
Architecturally, we design a mask decoder to generate class-agnostic masks directly from query images.
This contrasts with existing methods \cite{li2024visual, carion2025sam} that rely on prompt images,
which overemphasize prompt-query semantic alignment at the expense of mask quality.
Furthermore, our training loss explicitly accounts for novel query classes absent from the prompts.
This design simplifies training data sampling, promotes diverse training combinations, and crucially,
equips \ourmodel to recognize novel classes during inference.
Finally, we evaluate \ourmodel on 24 OOD benchmarks featuring distribution shifts, spanning natural, biological, ecological, cosmological, medical, and pathological imagery. 
Extensive experiments and analyses confirm \ourmodel's robust performance and its significantly superior performance over existing methods (\S\ref{sec:exp}).

We summarize three key contributions as follows:
\begin{enumerate}[leftmargin=10pt, topsep=-1pt, itemsep=3pt,parsep=-2pt]
    \item
    We introduce simple techniques, including  architecture design, training batch sampling strategy, and loss formulation, that enable training over heterogeneous datasets and efficient inference.
    
    \item 
    With these techniques, we train the foundation model \ourmodel and release it alongside source code to foster future research.

    \item 
    We establish a rigorous evaluation suite comprising 24 benchmarks from diverse domains
    and show that \ourmodel significantly outperforms existing approaches across these benchmarks.
    
\end{enumerate}

\section{Related work}
\label{sec:related-work}

\textbf{Vision Foundation Models (VFMs)} are trained on massive images, optionally along with other data modalities.
Some VFMs serve as backbones for encoding images.
For example,
MoCo models~\cite{he2020momentum, chen2021mocov3} learn transferable features via contrastive learning over unlabeled images;
MAE~\cite{he2022masked} and PixIO~\cite{yang2026pursuit} train VFMs 
with an objective of randomly masked region reconstruction;
DINO models \cite{oquab2023dinov2, simeoni2025dinov3} are trained further with self-distillation~\cite{caron2021emerging}.
Some others are trained over multimodal data. 
For example,
CLIP~\cite{radford2021learning}, 
MetaCLIP~\cite{xu2024demystifying, chuang2026meta},
SigLiP~\cite{zhai2023sigmoid},  and PE~\cite{bolya2026perception} train image and text encoders via multimodal contrastive learning on massive image-text pairs~\cite{schuhmann2022laion, xu2024demystifying, sun2024alpha},
producing visual-textual aligned features that allow open-vocabulary visual recognition.
In this work, 
we focus on training a VFM for in-context segmentation,
a more concrete application than learning image encoders and a mid-level vision problem compared with the high-level open-vocabulary recognition.

\begin{figure}[t]
    \centering
    \vspace{-2mm}
        \includegraphics[width=0.999\textwidth]{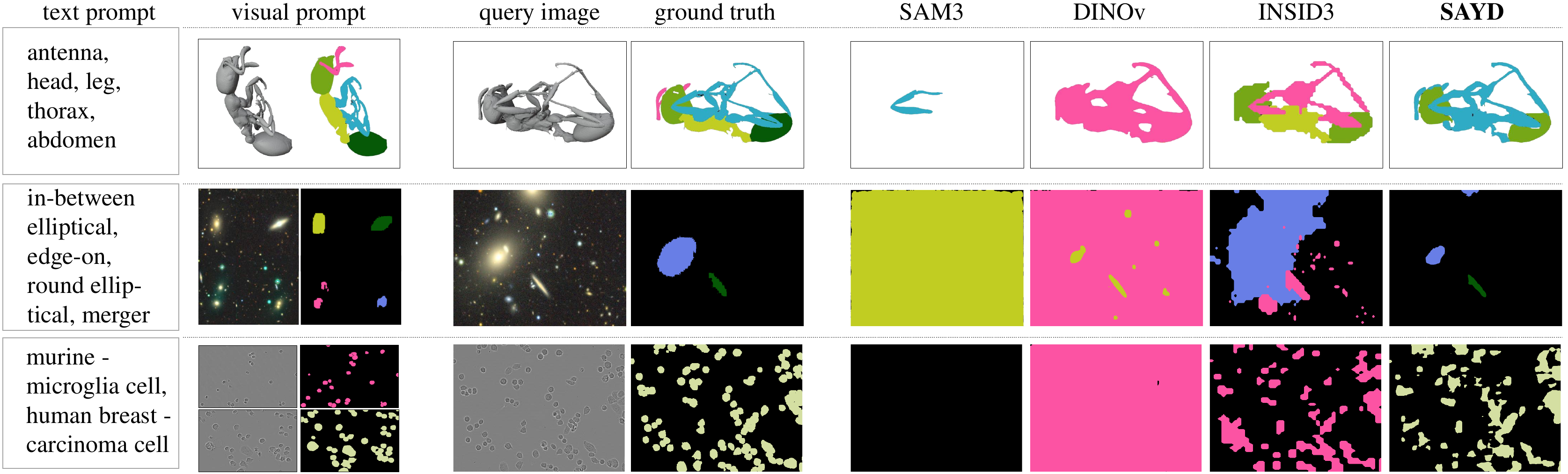}
\vspace{-5mm}
\caption{
\small
{\bf Qualitative comparison of segmentation models} through three examples.
SAM3 \cite{carion2025sam}, a prominent vision foundation model (VFM) for  open-vocabulary segmentation, fails to segment the targets using text prompts. 
DINOv \cite{li2024visual} and INSID3 \cite{cuttano2026insid3},
representative ICS models,
also struggle to accurately segment the targets.
In contrast, our \ourmodel, 
a VFM specifically designed for ICS, consistently produces the most accurate and visually coherent segmentation results.
}
\vspace{-1mm}
\label{fig:quick-visual-comparison} 
\end{figure}

\textbf{VFMs for image segmentation} can be broadly categorized by the type of prompt they accept.
First, \textit{high-level text prompts} (e.g., class names or descriptions \cite{fang2019modularized})
enable open-vocabulary segmentation, allowing users to segment images based on texts \cite{wei2023ov, ma2026autoexpert}.
Representative models include
X-Decoder~\cite{zou2023generalized}, OpenSeeD~\cite{zhang2023simple},
and SAM3~\cite{carion2025sam}.
However, they struggle in highly-specialized domains (Fig.~\ref{fig:quick-visual-comparison}), 
where target concepts are often associated with domain-specific terminology sparsely represented in the models' pretraining data or rely on tacit visual knowledge difficult to articulate unambiguously in language.
Second, \textit{low-level visual prompts} (e.g., points, boxes and masks) specify spatial locations, as in SAM \cite{kirillov2023segment}.
While well suited to interactive segmentation, they require manual prompts for each query image and are thus limited  in high-throughput applications.
Third, \textit{mid-level visual prompts} (i.e., images and their segmentation masks) specify the target visual concepts and instruct models to segment analogous regions in query images.
This ICS paradigm is particularly valuable for applications that rely on tacit visual knowledge, such as 
clinical judgement and medical practice \cite{patel1999expertise, thornton2006tacit} and
astronomical discovery \cite{biagioli1995tacit}.
We train a VFM with ICS generalization across diverse scientific domains.

{\bf In-Context Segmentation (ICS)}
adapts in-context learning from natural language processing \cite{brown2020language} to computer vision \cite{bar2022visual} and has gained increasing attention.
Early proxy-model approaches predict colored masks for query images \cite{wang2023images, wang2023seggpt} and thus are not optimized for ICS.
Other methods, such as Hummingbird \cite{balazevic2023towards} 
and INSID3 \cite{cuttano2026insid3},
use similarity metrics to associate off-the-shelf features between prompt and query images extracted by pretrained image encoders.
However, they are limited as the encoders are not optimized for ICS (Fig.~\ref{fig:pca}).
Furthermore,
DINOv \cite{li2024visual} is a universal visual in-context prompting model trained on 11M images sourced from the COCO \cite{lin2014microsoft} and SA-1B \cite{kirillov2023segment} datasets.
It supports not only ICS but also other prompt types (e.g., strokes, boxes, and points).
However,
its  universality comes at the expense of limited ICS ability.
Other existing methods face practical limitations:
Hummingbird \cite{balazevic2023towards} is not open-source,
while Matcher \cite{liu2023matcher}, SegIC \cite{meng2024segic}, and INSID3~\cite{cuttano2026insid3} produce only binary masks for single target concept, requiring multiple forward passes to segment multiple concepts and incurring substantial computational overhead.
We address these limitations by training a VFM specifically designed for efficient, robust, and cross-domain generalizable ICS.

\section{Problem Definition and Evaluation Protocol}
\label{sec:problem}

\textbf{Problem definition.} 
A VFM designed for ICS aims to segment coherent regions in a query image, where the regions are specified by visual demos consisting of prompt images and their corresponding segmentation masks (Fig.~\ref{fig:splashing_figure}).
This VFM should generalize to diverse and previously unseen domains without task-specific adaptation.
Therefore, the core problem is to develop such a VFM that enables robust, cross-domain generalizable, efficient learning-free ICS.

\textbf{Evaluation protocol.}
To comprehensively evaluate the foundational capability of the designed models,
we evaluate them on test data sourced from diverse unseen domains, e.g., from scientific imagery.
We further include a broad range of segmentation tasks, including instance, semantic, and panoptic segmentation,
across multiple levels of granularity, from object- to part-level segmentation.

\section{Methodology}
\label{sec:method}

Training a foundation model for ICS should be done over large-scale, diverse data with segmentation annotations spanning multiple levels of granularity.
Such data can be easily sourced from diverse public  datasets originally curated for various segmentation tasks.
However, training over such heterogeneous datasets has been underexplored and requires effective techniques.
We introduce such techniques regarding
architecture design,
training data sampling, and loss formulation.

\begin{figure}[t]
\centering
\vspace{-1.5mm}
\includegraphics[width=0.999\textwidth]{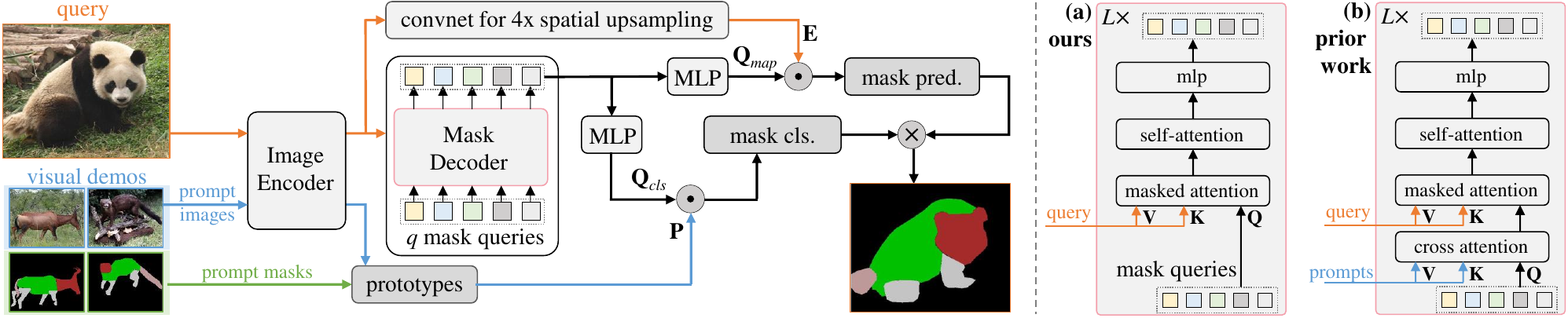}
\vspace{-5mm}
\caption{
\small
{\bf Left:} 
In \ourmodel, an image encoder extracts feature maps for the query and prompt images.
A mask decoder processes the query's feature map and outputs two sets of tokens (with transformations by MLPs):
$\mathbf{Q}_{map}$ for mask generation, and $\mathbf{Q}_{cls}$ for mask classification.
Meanwhile, the feature maps of prompt images are averaged into class prototypes $\mathbf{P}$ based on the prompt masks.
\ourmodel uses the prototypes to classify the tokens $\mathbf{Q}_{cls}$.
Combining the masks and their predicted classes produces segmentation results at the instance, semantic, and panoptic levels.
{\bf Right:} 
Our decoder design \textbf{(a)} takes as input only the query's feature map that generates \textit{class-agnostic} masks.
Differently,
the design \textbf{(b)}, widely adopted in prior work \cite{carion2025sam, li2024visual},
additionally requires prompts to generate \textit{class-aware} masks,
which overemphasize prompt-query semantic alignment at the expense of mask quality.
Our design not only reduces computational overhead but also improves ICS performance (Tables~\ref{tab:inference-latency}\&\ref{tab:decoder}).
}
\vspace{-1mm}
\label{fig:architecture}
\end{figure}

\subsection{Model Architecture}
\label{ssec:arch}

From first principles, 
we design the architecture of \ourmodel (Fig.~\ref{fig:architecture})
to achieve inference efficiency,
efficacy to support multiple prompt images,
and ability to produce high-quality masks.
\ourmodel comprises two main components, an image encoder and a lightweight mask decoder.
We describe them below, deferring implementation details to the Appendix \S\ref{sec_suppl:arch} and the supplemented code.

{\bf An image encoder} extracts feature maps for any input images.
For a query image in resolution $H\times W$,
it uses a convnet to spatially upsample the feature map and outputs $\mathbf{E}\in \mathbb{R}^{h \times w \times d}$.
The encoder architecture follows  Vision Transformer (ViT)~\cite{dosovitskiy2020image} and is initialized from pretrained weights (Table~\ref{tab:img-enc}).
We further incorporate window attention to improve computational efficiency \cite{li2022exploring} and apply a linear layer to reduce the feature dimension to $d$.
For visual demos supposedly defining $K$ classes,
the encoder extracts the prompt images' feature maps,
which are then spatially averaged for each of the $K$ classes, within relevant regions according to the prompt masks, resulting in $K$ prototypes $\mathbf{P} \in \mathbb{R}^{K \times d}$ (with L2 normalization).
The query's feature map $\mathbf{E}$ and the prototypes $\mathbf{P}$ are used for mask prediction and classification by the following decoder.

{\bf A lightweight mask decoder} maps the query's feature map $\mathbf{E}$ to pairs of segmentation masks and classes \cite{cheng2021per}.
Concretely,
with $q$ learnable mask queries,
it first maps $\mathbf{E}$ to $q$ tokens,
which are then transformed by two MLPs to two sets of L2-normalized tokens,
denoted as 
$\mathbf{Q}_{map} \in \mathbb{R}^{q\times d}$ and
$\mathbf{Q}_{cls} \in \mathbb{R}^{q \times d}$. They are used for mask prediction and classification, respectively:
\begin{itemize}[leftmargin=10pt, topsep=-1pt, itemsep=3pt,parsep=-2pt]

\item 
    {\em Mask prediction.} 
    With the query feature map $\mathbf{E}$ and tokens $\mathbf{Q}_{map}$,
    it computes $q$ masks $\mathbf{M} = \mathbf{E} \times_{3} \mathbf{Q}_{map}^T \in \mathbb{R}^{h \times  w \times q}$, which are then converted into binary masks.
    These masks are \textit{class-agnostic} \cite{kong2018recurrent} that depend solely on the query  (Fig.~\ref{fig:architecture}a).
    In contrast, existing methods  \cite{li2024visual, carion2025sam} additionally require prompt image features (Fig.~\ref{fig:architecture}b) and generate \textit{class-aware} masks.
    Consequently, they overemphasize prompt-query semantic alignment and degrade mask quality.
    Our design not only improves inference efficiency even with multiple prompts, but also facilitates the training of high-quality mask generation, thereby improving ICS  (Tables~\ref{tab:inference-latency}\&\ref{tab:decoder}).

\item
    {\em Mask classification.} 
    With $q$ tokens representing $q$ segment masks,
    we use prototypes $\mathbf{P}$ to compute per-token logits
    $\mathbf{S} = \mathbf{Q}_{cls}  \mathbf{P}^T \in \mathbb{R}^{q\times K}$.
    Straightforwardly, one could apply argmax to obtain the classification results for these tokens.
    However,
    the query image likely contains novel classes absent from the prompt images (and therefore not represented by the prototypes).
    We address this by training \ourmodel to identify the novel classes, detailed in the following two subsections.

\end{itemize}

\subsection{Training Data Sampling}
\label{ssec:sampling}

We repurpose 12  publicly-available datasets (listed in Appendix Table~\ref{tab:appendix-dataset}), originally curated for panoptic segmentation on diverse images, for training \ourmodel.
As different datasets name visual concepts differently,
we adopt a within-dataset training data sampling  strategy.

{\bf Prompt-query sampling}.
We sample training data within each  dataset.
Specifically, for a given dataset, 
we randomly sample one image as the query and $N$ additional images, along with their ground-truth masks,
as visual demos.
The ground-truth masks of both query and prompt images help identify regions that share class labels to compute training loss (detailed in \S\ref{ssec:loss}).
Despite the simplicity of this sampling strategy,
a recurring issue arise:
the query image contains novel classes absent from the prompt images.
We use the term \textit{ignota} to denote these novel classes present in the query but missing from the visual demos.

{\bf The \textit{ignota}}. 
Existing ICS methods that train models circumvent \textit{ignota} in various ways.
Some train regression models to color and group pixels for segmentation \cite{wang2023seggpt, wang2023images}. This sidesteps the problem entirely but the resulting models are not optimized for ICS.
\cite{meng2024segic} design a model to handle only one class per run.
This simplifies the sampling of query and prompt images (requiring only one shared class) but renders training inefficiency as each iteration ignores all other labeled classes.
\cite{liu2024simple} restrict their sampling to cases where all query classes are present in the prompts \cite{SINE_github_sampling}.
This severely limits the possible prompt-query combinations.
\cite{li2024visual} use the current training batch as prompts and treat each individual example within the batch as a query.
While this eliminates the \textit{ignota} issue during training,
it simultaneously introduces shortcut learning and prevents learning of a model capable of identifying \textit{ignota} at test time.
In the real world,
numerous applications solicit
the ability to identify novel classes \cite{scheirer2012toward, kong2021opengan, adaime2024deep, chakravarthy2025lidar}.
Hence,
we explicitly incorporate \textit{ignota} into the training loss computation (detailed next).
Empirically, doing so not only facilitates training but also yields superior ICS performance  (Table~\ref{tab:ignota}).

\subsection{Loss Function}
\label{ssec:loss}

By design, 
\ourmodel produces mask queries, from which it generates class-agnostic masks and class labels (Fig.~\ref{fig:architecture}).
We train \ourmodel using task-specific losses.
For mask prediction,
we employ the widely used dice loss $\mathcal{L}_{\text{dice}}$ and focal loss $\mathcal{L}_{\text{focal}}$~\cite{cheng2021per}. 
Our novelty of loss formulation lies in mask classification,
detailed below.

\textbf{Loss for mask classification.}
Recall that \ourmodel computes $q$ per-token logits $\mathbf{S} \in \mathbb{R}^{q\times K}$ (\S\ref{ssec:arch}),
to which a cross-entropy (CE) loss could be straightforwardly applied.
However, the training data contain \textit{ignota}.
When using a CE loss, there are two ways to handle \textit{ignota}:
(1) ignoring them during loss computation,
or (2) assigning them to an additional placeholder class.
The former forgoes the opportunity of exploiting \textit{ignota} to train models capable of identifying novel classes, while the latter inevitably collapses diverse \textit{ignota} into a single class and thus complicates training.
Hence, we instead employ the binary cross-entropy (BCE) loss $\mathcal{L}_{\text{bce}}$.
Under the BCE formulation,
the features of \textit{ignota}, as well as background regions (cf. the black regions in the fourth column of Fig.~\ref{fig:splashing_figure}), 
serve as negative examples for all $K$ prototypes.
Furthermore, 
we learn a temperature $\tau$ and a bias $\beta$ to scale and calibrate the logits, i.e.,
$\tilde{\mathbf{S}} = \frac{\mathbf{S}}{\tau} + \beta \in \mathbb{R}^{q\times K}$, before passing them to the BCE loss.
Learning both $\tau$ and $\beta$ is crucial for  successfully training \ourmodel, as shown in Table~\ref{tab:bias}.

\textbf{Remarks.}
The adoption of BCE loss with learnable temperature and bias is a novel contribution to the ICS literature.
As discussed in \S\ref{ssec:sampling},
prior methods circumvent \textit{ignota} during training,
thereby avoiding the need to compute a loss for them.
That said, they assume  no \textit{ignota},
preventing training models to identify novel classes at test time.
Instead,  
learning with \textit{ignota} yields superior ICS models (Tables \ref{tab:ignota}\&\ref{tab:loss}).
Moreover,
DINOv \cite{li2024visual} employs a BCE focal loss \textit{without} a learnable bias and eschews L2-normalization when computing similarities between prototypes and query features.
Despite this alterative strategy for modulating feature similarities during loss computation,
incorporating a learnable bias significantly improves model performance (Table~\ref{tab:bias}).

\textbf{The final loss.}
Combining the aforementioned loss terms, we train \ourmodel using the final loss
$\mathcal{L} = 20 \cdot \mathcal{L}_{\text{bce}} + 2 \cdot \mathcal{L}_{\text{focal}} + \mathcal{L}_{\text{dice}}$.
We set the weights of loss terms by balancing the magnitude of their produced gradient.
We provide details in Appendix \S\ref{sec_suppl:arch}.

\subsection{Inference}
\label{ssec:inference}

\ourmodel segments a query image conditioned on visual demos,
which define target classes.
It outputs class-agnostic binary masks, each capturing a potential object instance.
By classifying and merging the masks according to the target classes,
\ourmodel achieves semantic and panoptic segmentation.
Finally,
for each  mask,
if its maximum calibrated logit (in the corresponding row of $\tilde{\mathbf{S}}$) falls below zero,
the region of this mask is classified as \textit{ignota};
otherwise, it is assigned to the argmax class.

\section{Experiments}
\label{sec:exp}

We validate \ourmodel by comparing it against recent and representative VFMs across 24 benchmarks, supported by comprehensive analyses and ablation studies.
We begin by outlining the experimental setup and key implementation details.

\subsection{Setup and Implementation}
\label{ssec:implementation}

\textbf{Training data.}
We combine 12 public datasets designed for panoptic segmentation, including COCO~\cite{lin2014microsoft}, ADE20K~\cite{zhou2017scene}, and Cityscapes~\cite{Cordts2016Cityscapes}.
Together, they provide 0.91M images and 11.46M segmentation masks to train \ourmodel.
See the ``in-domain test examples'' in Fig.~\ref{fig:splashing_figure} for some images from these datasets,
and Appendix \S\ref{sec_suppl:datasets} for more details.

\textbf{Benchmark datasets.}
We use 24 public datasets span diverse domains, including gaming scenes~\cite{richter2016playing}, art paintings~\cite{cohen2022semantic},
object parts~\cite{gong2017look, mu2020learning, katzke2026high}, animal/plant species~\cite{wah2011caltech, wei2409plantseg, he2024sprout}, 
remote sensing~\cite{Wang2021LoveDA},
medical imaging~\cite{shiraishi2000development},
manufacturing materials~\cite{metaldam2024}, astronomy~\cite{willett2013galaxy}, and cell biology~\cite{edlund2021livecell}.
These datasets are  clearly different from the aforementioned 12 training datasets,
hence serving as OOD benchmarks.
Refer to the ``out-of-domain test examples'' in Fig.~\ref{fig:splashing_figure} for some test images and Appendix \S\ref{sec_suppl:datasets} for more details.
For each benchmark,
we randomly sample $\sim$10 images from its official training set as visual demos, ensuring that they collectively cover all semantic classes of this benchmark,
and use the official validation/test images as queries.
All queries within a benchmark share the same visual demos.
For each method on each benchmark,
we repeat the evaluation with three sets of independently sampled visual demos and report the average metrics.
We evaluate semantic, instance, and panoptic segmentation using mean Intersection-over-Union (mIoU), Average Precision (AP), and Panoptic Quality (PQ), respectively.

\textbf{Compared methods.}
We compare \ourmodel with methods from two prevailing paradigms:
text-based and vision-based prompting. 
We provide their detailed specifications in Appendix Table~\ref{tab:appendix-methods} and briefly describe them below:
\begin{itemize}[leftmargin=10pt, topsep=-1pt, itemsep=3pt,parsep=-2pt]

    \item \textit{Text-based prompting}. We compare leading open-vocabulary segmentation models, including X-Decoder (X-Dec)~\cite{zou2023generalized}, OpenSeeD (O-SeeD)~\cite{zhang2023simple}, APE~\cite{shen2024aligning},
    Grounded SAM (G-SAM)\cite{ren2024grounded},
    OpenWorldSAM (O-SAM) \cite{xiao2025openworldsam},
     RADSeg+SAM (R-SAM) \cite{alama2026radseg},
    and SAM3 \cite{carion2025sam}.
    They use textual descriptions of target concepts (e.g., semantic class names) as prompts.
    \revised{We also evaluate the recently released open-source Multimodal Large Language Model (MLLM) WeVisOne-VL-4B model~\cite{liang2026wevisone}.
    Following its official pipeline, we report performance under the following two types of text prompts: 
    (1) \textit{WeV-cls}, standing for only using class names (provided by official benchmarks) as shown in Fig.~\ref{fig:quick-visual-comparison} -left,
    and (2) \textit{WeV-ins}, which combines both the class names and corresponding one-sentence descriptions.
    Each description is obtained by prompting the AI agent Qwen3.8 \cite{qwen3.8} to describe a target class name.
    For example, for \textit{person}, the description is ``\textit{a human being, including any man, woman or child in the scene''.}
    }
    
    \item \textit{Vision-based prompting}. We compare against Painter~\cite{wang2023images}, \revised{SegGPT}~\cite{wang2023seggpt}, 
    \revised{UniverSeg}~\cite{butoi2023universeg},
    \revised{Matcher}~\cite{liu2023matcher},
    \revised{SINE}~\cite{liu2024simple},
    \revised{SegIC}~\cite{meng2024segic},
    DINOv~\cite{li2024visual}, INSID3~\cite{cuttano2026insid3},
    \revised{SAM2}~\cite{ravi2024sam2}, and
    \revised{SAM3-V which is a mode of SAM3 that supports images as prompts}.
    To the best of our knowledge,
    INSID3 is the most recent ICS method, which exploits a frozen  DINOv3 \cite{simeoni2025dinov3} as the image encoder,
    while DINOv represents the most recent approach to training an ICS model on large-scale data.
    \revised{Since Matcher, SegIC, and INSID3 can only generate a binary mask for one class per inference,
    we apply it sequentially to each class and combine the resulting binary masks to obtain multi-class semantic segmentation. 
    Moreover, since the weights of SegGPT are not publicly available, we retrain it  using the same data as our SAYD. 
    Notably, while supporting visuals in prompts,
    SAM3-V acutally requires low-level prompts such as boxes on the query image itself. To enable SAM3-V to operate as an ICS model,
    we concatenate the prompt images with the query into a composite image, where the regions consisting of prompt images contain such boxes.
    }

\end{itemize}

\begin{table}[t]
\centering
\caption{\small
    \textbf{Benchmark results for semantic segmentation} w.r.t mIoU. 
    SAM3 is the most competitive approach. It performs well on datasets of common objects (e.g., GTA) but significantly degrades on highly-specialized ones such as TigD, Plant, Ant, Gala, and Cell,
    where its in-house training data likely lack sufficient representation.
    By contrast, \ourmodel exhibits superior generalization to these domains.
}
\vspace{-3mm}
\scalebox{0.82}{
    \small
    \renewcommand{\arraystretch}{1.1}
    \setlength{\tabcolsep}{2.4pt}
    \begin{tabular}{l *{15}{c}}
    \toprule
        & \multicolumn{7}{c}{common object categories} & \multicolumn{8}{c}{uncommon concepts}\\
        \cmidrule(lr){2-8}
        \cmidrule(lr){9-16}
        Model & GTA & DRAM & CUB & LIP & YouT & NYU & iShape & TigD & Plant & Ant & LDA & JSRT & ODM & Gala & Cell\\ 
                
        \midrule
        \rowcolor{lightlightgrey} \multicolumn{16}{l}{\textit{Open-Vocabulary Segmentation}} \\
        
        X-Dec \tiny\cite{zou2023generalized} & 42.87 & 27.90 & \textcolor{white}{o}3.53 & \textcolor{white}{o}4.14 & 32.78 & 33.51 & 25.10 & \textcolor{white}{o}1.43 & \textcolor{white}{o}1.54 & \textcolor{white}{o}1.03 & 28.72 & \textcolor{white}{o}9.86 & \textcolor{white}{o}0.84 & \textcolor{white}{o}0.09 & \textcolor{white}{o}0.66\\

        O-SeeD \tiny\cite{zhang2023simple} & 41.21 & 25.93 & \textcolor{white}{o}0.11 & \textcolor{white}{o}1.82 & \textcolor{white}{o}7.95 & 31.04 & 11.70 & \textcolor{white}{o}0.19 & \textcolor{white}{o}1.09 & \textcolor{white}{o}5.25 & 13.02 & \textcolor{white}{o}4.85 & \textcolor{white}{o}1.23 & \textcolor{white}{o}0.01 & \textcolor{white}{o}1.90\\

        APE \tiny\cite{shen2024aligning} & 24.62 & 53.23 & \textcolor{white}{o}0.00 & 13.05 & 28.24 & \textbf{42.33} & 21.01 & \textcolor{white}{o}4.29 & \textcolor{white}{o}1.32 & \textcolor{white}{o}2.01 & 18.17 & \textcolor{white}{o}0.64 & \textcolor{white}{o}0.74 & \textcolor{white}{o}0.03 & \textcolor{white}{o}0.37\\
        
        G-SAM \tiny\cite{ren2024grounded} & 42.88 & 42.08 & \textcolor{white}{o}0.02 & 16.95 & 14.15 & 29.45 & \textcolor{white}{o}5.45 & \textcolor{white}{o}7.89 & \textcolor{white}{o}0.04 & \textcolor{white}{o}8.78 & 10.84 & 11.34 & \textcolor{white}{o}0.22 & \textcolor{white}{o}3.12 & \textcolor{white}{o}0.43\\
        
        O-SAM \tiny\cite{xiao2025openworldsam} & 28.50 & 30.40 & \textcolor{white}{o}0.12 & \textcolor{white}{o}8.45 & \textcolor{white}{o}4.75 & 27.84 & \textcolor{white}{o}6.05 & \textcolor{white}{o}2.99 & \textcolor{white}{o}0.19 & \textcolor{white}{o}5.34 & 26.37 & 10.70 & \textcolor{white}{o}4.83 & \textcolor{white}{o}1.01 & \textcolor{white}{o}0.80\\
        
        R-SAM \tiny\cite{alama2026radseg} & 37.35 & 31.42 & \textcolor{white}{o}6.58 & 20.16 & 22.57 & 35.90 & 31.12 & \textcolor{white}{o}0.96 & \textcolor{white}{o}2.37 & \textcolor{white}{o}2.66 & 25.00 & 10.57 & \textcolor{white}{o}2.56 & \textcolor{white}{o}1.44 & \textcolor{white}{o}0.00\\

        \revised{WeV-cls} \tiny\cite{liang2026wevisone} & 39.27 & 49.45 & \textcolor{white}{o}1.53 & 14.40 & \textcolor{white}{o}6.76 & 21.26 & \textcolor{white}{o}8.98 & 14.05 & \textcolor{white}{o}0.13 & 21.86 & 21.73 & 13.80 & \textcolor{white}{o}1.97 & \textcolor{white}{o}0.32 & \textcolor{white}{o}3.01\\

        \revised{WeV-ins} \tiny\cite{liang2026wevisone} & 38.62 & 58.33 & \textcolor{white}{o}2.76 & 17.74 & \textcolor{white}{o}6.21 & 19.78 & \textcolor{white}{o}8.91 & 14.76 & \textcolor{white}{o}1.12 & 31.09 & 24.52 & \textcolor{white}{o}9.08 & \textcolor{white}{o}1.64 & \textcolor{white}{o}0.41 & \textcolor{white}{o}0.96\\
        
        SAM3 \tiny\cite{carion2025sam} & \textbf{54.34} & \textbf{67.62} & \textbf{24.02} & \textbf{33.89} & 55.97 & 38.78 & 13.15 & 13.51 & \textcolor{white}{o}0.64 & \textcolor{white}{o}9.36 & 30.49 & \textcolor{white}{o}7.93 & \textcolor{white}{o}0.00 & \textcolor{white}{o}1.35 & \textcolor{white}{o}0.00\\

        \midrule
        \rowcolor{lightlightgrey} \multicolumn{16}{l}{\textit{In-Context Segmentation}} \\
        
        Painter \tiny\cite{wang2023images} & \textcolor{white}{o}0.16 & \textcolor{white}{o}0.27 & \textcolor{white}{o}0.00 & \textcolor{white}{o}0.10 & \textcolor{white}{o}0.09 & \textcolor{white}{o}2.35 & \textcolor{white}{o}0.82 & \textcolor{white}{o}0.68 & \textcolor{white}{o}0.03 & \textcolor{white}{o}1.96 & \textcolor{white}{o}5.20 & 12.06 & \textcolor{white}{o}0.55 & \textcolor{white}{o}0.34 & \textcolor{white}{o}0.67\\

        \revised{SegGPT} \tiny\cite{wang2023seggpt} & \textcolor{white}{o}1.77 & \textcolor{white}{o}0.42 & \textcolor{white}{o}0.06 & \textcolor{white}{o}1.24 & \textcolor{white}{o}0.24 & \textcolor{white}{o}0.78 & \textcolor{white}{o}4.60 & \textcolor{white}{o}1.09 & \textcolor{white}{o}0.02 & \textcolor{white}{o}4.83 & \textcolor{white}{o}7.61 & \textcolor{white}{o}9.88 & \textcolor{white}{o}0.96 & \textcolor{white}{o}0.78 & \textcolor{white}{o}1.24\\

        \revised{UniverSeg} \tiny\cite{butoi2023universeg} & \textcolor{white}{o}0.03 & \textcolor{white}{o}0.18 & \textcolor{white}{o}0.00 & \textcolor{white}{o}0.62 & \textcolor{white}{o}0.14 & \textcolor{white}{o}0.28 & \textcolor{white}{o}1.45 & \textcolor{white}{o}0.00 & \textcolor{white}{o}0.00 & \textcolor{white}{o}2.22 & \textcolor{white}{o}1.05 & \textcolor{white}{o}2.92 & \textcolor{white}{o}0.11 & \textcolor{white}{o}0.37 & \textcolor{white}{o}1.97\\

        \revised{Matcher} \tiny\cite{liu2023matcher} & 41.51 & 46.98 & 17.36 & 10.95 & 30.06 & 33.88 & 22.68 & 10.86 & \textcolor{white}{o}9.56 & 27.90 & 23.16 & 14.03 & 17.46 & \textcolor{white}{o}1.45 & 13.87\\

        \revised{SINE} \tiny\cite{liu2024simple} & 36.49 & 39.23 & \textcolor{white}{o}0.07 & \textcolor{white}{o}6.93 & 10.40 & 32.37 & 24.71 & \textcolor{white}{o}4.73 & \textcolor{white}{o}0.89 & 25.97 & 31.52 & \textcolor{white}{o}0.18 & 13.73 & \textcolor{white}{o}1.52 & \textcolor{white}{o}2.30\\

        \revised{SegIC} \tiny\cite{meng2024segic} & 32.54 & 31.22 & \textcolor{white}{o}1.32 & 13.68 & \textcolor{white}{o}7.90 & 29.02 & 27.44 & 12.58 & \textcolor{white}{o}1.65 & 26.22 & 33.72 & \textcolor{white}{o}0.33 & 18.57 & \textcolor{white}{o}5.72 & \textcolor{white}{o}1.58\\

        \revised{SAM2} \tiny\cite{ravi2024sam2} & \textcolor{white}{o}0.73 & \textcolor{white}{o}2.14 & \textcolor{white}{o}0.81 & \textcolor{white}{o}4.56 & \textcolor{white}{o}0.58 & \textcolor{white}{o}1.12 & \textcolor{white}{o}1.88 & \textcolor{white}{o}0.03 & \textcolor{white}{o}2.38 & \textcolor{white}{o}8.45 & \textcolor{white}{o}1.91 & \textcolor{white}{o}2.97 & \textcolor{white}{o}3.37 & \textcolor{white}{o}0.57 & \textcolor{white}{o}2.04\\
                
        DINOv \tiny\cite{li2024visual} & 16.37 & 27.42 & \textcolor{white}{o}4.03 & \textcolor{white}{o}5.54 & 11.91 & 30.48 & 22.47 & \textcolor{white}{o}1.88 & \textcolor{white}{o}0.94 & 15.65 & 16.45 & 15.40 & \textbf{18.88} & \textcolor{white}{o}1.70 & \textcolor{white}{o}5.28\\
        
        INSID3 \tiny\cite{cuttano2026insid3} & 15.28 & 18.35 & \textcolor{white}{o}0.00 & \textcolor{white}{o}7.32 & \textcolor{white}{o}9.86 & 10.68 & 21.97 & \textcolor{white}{o}5.59 & \textcolor{white}{o}0.03 & 19.85 & 31.15 & 10.52 & 16.39 & \textcolor{white}{o}1.69 & \textcolor{white}{o}2.63\\

        \revised{SAM3-V} \tiny\cite{carion2025sam} & \textcolor{white}{o}4.81 & \textcolor{white}{o}3.47 & 17.82 & 20.09 & \textcolor{white}{o}2.91 & \textcolor{white}{o}3.84 & \textcolor{white}{o}5.87 & \textcolor{white}{o}1.77 & \textcolor{white}{o}4.03 & 10.37 & \textcolor{white}{o}1.12 & \textcolor{white}{o}6.49 & \textcolor{white}{o}5.59 & \textcolor{white}{o}1.28 & \textcolor{white}{o}1.89\\
        
        \rowcolor{col1} 
        \ourmodel (ours) & 52.95 & 58.07 & 17.06 & 29.65 & \textbf{58.51} & 40.87 & \textbf{34.92} & \textbf{15.52} & \textbf{15.62} & \textbf{32.04} & \textbf{39.91} & \textbf{16.88} & 18.73 & \textbf{10.19} & \textbf{22.43}\\
        \bottomrule
    \end{tabular}
}
\vspace{-0mm}
\label{tab:sota-miou}
\end{table}

{\bf Implementation.}
For benchmarking, 
we initialize the image encoder with pretrained DINOv3 ViT-L~\cite{simeoni2025dinov3, dosovitskiy2020image};
we validate  \ourmodel with other pretrained encoders in Table~\ref{tab:img-enc}.
The mask decoder has 4 Transformer blocks and processes $q=200$ 256d mask queries.
We use random horizontal flipping and random scale jittering in the range [0.5, 2.0] for data augmentation.
We train \ourmodel for 550k iterations using the AdamW optimizer~\cite{loshchilov2017decoupled} with step-wise learning rate decay.
The learning rate is linearly warmed up~\cite{goyal2017accurate} to 3e-4 in the first 10k iterations and reduced by 10$\times$ at iteration 500k.
For the image encoder, we use a layer-wise learning-rate decay~\cite{bao2021beit} of 0.75 and a drop-path rate of 0.3. 
We apply deep supervision~\cite{cheng2021per} with training losses in all decoder blocks. 
Each training batch contains 128 samples:
each sample consists of $N=5$ prompt images and 1 query image, all at 512$\times$512 pixel resolution.

\subsection{Results}
\label{ssec:benchmarking}

Table~\ref{tab:sota-miou} benchmarks methods on 15 semantic segmentation benchmarks.
SAM3, the most recent open-vocabulary segmentation model, 
works well on benchmarks that focus on common objects (e.g., GTA, DRAM, CUB and LIP) but struggles on scientific domains,
which specify target concepts using domain-specific terminology or preferably tacit visual knowledge (e.g., ODM~\cite{metaldam2024}, Gala~\cite{willett2013galaxy}, and Cell~\cite{edlund2021livecell}).
Refer to Fig.~\ref{fig:quick-visual-comparison} for its segmentation results on some images from these domains.
\revised{Regarding the most recent MLLM WeVisOne-VL-4B,
its WeV-ins mode improves over WeV-cls by incorporating descriptions.
Yet, similar to SAM3, it works well on benchmarks focusing on common objects but still struggle on the rest.}
Our \ourmodel rivals SAM3 and WeVisOne-VL-4B on the benchmarks of common objects and significantly outperforms them on scientific ones.
Notably, \ourmodel is trained on $<$1M images from public datasets, whereas SAM3 is trained over 5.2M proprietary real images plus much more video frames and synthetic data.

The compared ICS methods notably underperform \ourmodel.
Painter, an early method that learns to regress/color pixels, 
is not directly designed for ICS and achieves poor performance.
INSID3, relying on the frozen DINOv3 image encoder which is not pretrained to solve ICS, exhibits limited performance.
DINOv trains a universal visual in-context prompting model that supports not only ICS but also other prompts (e.g., points and boxes).
It trains on COCO with mask classification loss, and SA-1B for additional mask generation loss (without class labels).
Its training dataset contains $>$11M images, compared with ours 0.9M images.
However, it still remarkably underperforms our \ourmodel, presumably because its focus on universality rather than optimizing the ICS function.
\revised{UniverSeg, a VFM for medical image segmentation, performs poorly,
likely due to that it accepts grayscale images only while the benchmark images are RGB.}

Fig.~\ref{fig:pca} qualitatively compares the representation of image encoders trained via different methods.
The visuals demonstrate that our \ourmodel learns features that helps ICS at different levels of granularity.
Table~\ref{tab:sota-ins} compares  \ourmodel, SAM3 and DINOv on 9 more datasets:
5 for instance segmentation and 4 for panoptic segmentation.
Note that INSID3 is designed only for in-context (binary) \textit{semantic} segmentation and cannot output per-instance masks,
hence we exclude it in these benchmarks.
In short, the comparisons demonstrate the superior performance of \ourmodel.

\begin{table}[t]
\centering
\caption{\small
\textbf{Benchmark results for instance and panoptic segmentation} w.r.t AP and PQ, respectively.
\ourmodel significantly outperforms DINOv and achieves performance superior or comparable to SAM3 across most datasets.
The only exception is the Fashionpedia benchmark, which contain apparel categories, likely better represented in SAM3's pretraining data.
}
\vspace{-3.5mm}
\scalebox{0.87}{
    \small
    \renewcommand{\arraystretch}{1.1}
    \setlength{\tabcolsep}{3pt}
    \begin{tabular}{l *{9}{c}}
    \toprule
        & \multicolumn{5}{c}{Instance Segmentation} & \multicolumn{4}{c}{Panoptic Segmentation}\\
        \cmidrule(lr){2-6}
        \cmidrule(lr){7-10}
        Model & PhenoBench & Fashionpedia & iSAID & ScanNet & TrashCan & WildDash2 & BSB & PASTIS & Hypersim\\  
        
        \midrule
        DINOv \tiny\cite{li2024visual} & \textcolor{white}{o}9.26 & 12.45 & \textcolor{white}{o}4.28 & \textcolor{white}{o}1.29 & \textcolor{white}{o}0.55 & \textcolor{white}{o}3.51 & 10.16 & \textcolor{white}{o}0.78 & 26.40\\
        SAM3 \tiny\cite{carion2025sam} & \textbf{18.71} & \textbf{34.93} & 11.56 & \textcolor{white}{o}9.75 & 11.72 & \textbf{16.53} & 21.60 & 13.61 & 43.25\\
        
        \rowcolor{col1}\ourmodel (ours) & 16.28 & 27.60 & \textbf{15.44} & \textbf{13.56} & \textbf{14.03} & 16.30 & \textbf{22.89} & \textbf{19.52} & \textbf{53.30}\\
        \bottomrule
    \end{tabular}
}
\vspace{-0mm}
\label{tab:sota-ins}
\end{table}

\begin{table}[t]
\centering
\caption{\small 
    {\bf Comparison of computational costs}. 
    \textbf{Left:} 
    We report model parameters, peak GPU memory, and FLOPs (computed via \texttt{fvcore}) for processing one query image with one prompt image.
    ``Image runs'' and ``Prompt runs'' indicate the number of forward passes required to predict  $K$ classes defined by the prompt.
    The right panel plots the wall-clock time (in seconds) versus class number $K$.
    With the simplified mask decoder (Fig.~\ref{fig:architecture}a),
    our \ourmodel maintains a constantly low computational cost, whereas others induce linear computing overheads w.r.t $K$.
}
\vspace{-3mm}
\begin{minipage}{0.48\textwidth}
\small 
\vspace{-2mm}
\scalebox{0.85}{
    \setlength{\tabcolsep}{1.5pt}
    \renewcommand{\arraystretch}{1.1}
        \begin{tabular}{lrrrrr}
            \toprule
            \multirow{2}{*}{Model} & \#Params & Memory & FLOPs & Image & Prompt \\
            & (M)  & (GB)  & (T)   & runs  & runs \\
            \midrule
            WeVisOne \tiny \cite{liang2026wevisone} & 4100.0 & 22.6 & 14.5 & $K$ & $K$ \\
            SAM3 \tiny \cite{carion2025sam} & 840.5 & 4.0 & 1.8 & 1 & $K$ \\
            DINOv \tiny \cite{li2024visual} & 224.6 & 3.4 & 1.6 & $K$ & $K$ \\
            INSID3 \tiny \cite{cuttano2026insid3} & 303.2 & 3.4 & 2.5 & $K$ & $K$ \\
            \rowcolor{col1} \ourmodel & 310.7 & 1.6 & 0.3 & 1 & 1 \\
            \bottomrule
        \end{tabular}
}
\end{minipage}
\medskip
\hfill
\hspace{1mm}
\begin{minipage}{0.46\textwidth}
    \centering
    \includegraphics[width=0.98\linewidth]{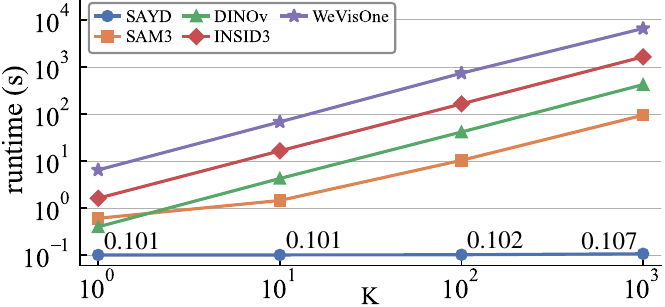}
\end{minipage}
\label{tab:inference-latency}
\vspace{-3mm}
\end{table}

\subsection{Analyses and Ablation Studies}
\label{ssec:ablation}

We analyze \ourmodel w.r.t inference efficiency
and ablate its key components.
For all analyses in this section,
we train \ourmodel for 100k iterations and use DRAM~\cite{cohen2022semantic}, TigD~\cite{mu2020learning}, \revised{Ant}~\cite{katzke2026high}, \revised{LDA}~\cite{Wang2021LoveDA}, and \revised{Gala}~\cite{willett2013galaxy} benchmarks to report results.

{\bf Inference efficiency.} 
Table~\ref{tab:inference-latency} compares the computational costs of our \ourmodel and representative methods.
Recall that \ourmodel adopts a lightweight mask decoder that takes as input only the query's features without conditioning on prompts  (Fig.~\ref{fig:architecture}a).
This supports pre-computing prototypes defined by visual demos and consequently helps \ourmodel maintain a constant runtime regardless of the number of classes.
In contrast, existing decoders either iteratively process each prompt atop the query's features (e.g., SAM3) or require a joint query-prompt forward pass for each individual class (e.g., DINOv and INSID3; Fig.~\ref{fig:architecture}b). 
Consequently, their inference time scales linearly with the number of classes.
Next, we compare architectural designs.

\begin{table}[t]
\centering
\caption{\small
    \revised{\textbf{Comparison of decoder designs.}
    We validate our proposed decoder design (Fig.~\ref{fig:architecture}a) against  prevailing alternatives (Fig.~\ref{fig:architecture}b), which rely on either the prompt images (\textbf{b-1}) or class prototypes (\textbf{b-2}) for mask prediction.
    \textbf{Left:} 
    We compare these designs w.r.t parameter count,
    FLOPS, and segmentation performance across benchmarks,
    highlighting the efficiency and effectiveness of our design.
    \textbf{Right:}
    Existing decoders require prompt images or class prototypes to produce class-aware masks, which can overemphasize prompt-query semantic alignment at the expense of fine-grained mask quality.
    We validate this by evaluating boundary segmentation quality using Boundary SQ, alongside standard recognition quality (RQ) and segmentation quality (SQ) metrics.
    Results on the Hypersim panoptic segmentation benchmark clearly demonstrate this point.}
}
\vspace{-3.3mm}
\begin{minipage}{0.56\textwidth}
\centering
\scalebox{0.8}{
    \setlength{\tabcolsep}{4pt}
    \renewcommand{\arraystretch}{1.1}

        \begin{tabular}{l *{7}{c}}
            \toprule
            arch. & Params$\downarrow$ & FLOPs$\downarrow$ & DRAM$\uparrow$ & TigD$\uparrow$ & \revised{Ant$\uparrow$} & \revised{LDA$\uparrow$} & \revised{Gala$\uparrow$}\\
            
            \midrule
            
            \rowcolor{col1} \textbf{a (ours)} & \textbf{6.31M} & \textcolor{white}{o}\textbf{2.20G} & \textbf{59.20} & \textbf{12.69} & \textbf{31.76} & \textbf{39.85} & \textbf{10.54}\\
            
            \textbf{b-1} & 7.37M & 25.25G & 50.45 & \textcolor{white}{o}5.12 & 22.49 & 30.98 & \textcolor{white}{o}2.80\\
            
            \textbf{b-2} & 7.37M & \textcolor{white}{o}2.31G & 52.96 & 11.32 & 25.27 & 33.24 & \textcolor{white}{o}2.51\\

            \bottomrule
        \end{tabular}
}
\end{minipage}
\hfill
\begin{minipage}{0.4\textwidth}
\centering
\scalebox{0.8}{
    \setlength{\tabcolsep}{4pt}
    \renewcommand{\arraystretch}{1.1}

        \begin{tabular}{lccc}
            \toprule
            arch. & \revised{RQ$\uparrow$} & \revised{SQ$\uparrow$} & \revised{Boundary SQ$\uparrow$}\\
            \midrule
            \rowcolor{col1} \textbf{a (ours)} & \textbf{61.08} & \textbf{81.66} & \textbf{76.05}\\
            
            \textbf{b-1} & 59.42 & 64.91 & 57.94\\
            
            \textbf{b-2} & 60.17 & 70.62 & 62.58\\

            \bottomrule
        \end{tabular}
}
\end{minipage}
\vspace{-2mm}
\label{tab:decoder}
\end{table}

\textbf{Simplified  mask decoder design.}
\ourmodel's mask decoder takes only the query as input (Fig.~\ref{fig:architecture}a; 
Table~\ref{tab:decoder} design \textbf{a}),
unlike existing decoders that condition on prompt images,
by requiring either feature maps of prompt images (Table~\ref{tab:decoder} design \textbf{b-1})
or class prototypes computed on the prompt images (design \textbf{b-2}, as in DINOv).
Comparing four-block decoders of these designs atop the same image encoder confirms \ourmodel's computational efficiency w.r.t parameter count and FLOPs, as well as its superior performance.
We conjecture that conditioning on prompts biases the decoder toward coarse semantic regions at the expense of mask boundaries,
thereby degrading segmentation performance. 
\revised{To validate this hypothesis, we conduct a breakdown analysis 
by comparing these designs w.r.t metrics Recognition Quality (RQ), Segmentation Quality (SQ), and Boundary SQ on the Hypersim panoptic segmentation benchmark~\cite{roberts2021Photorealistic}. 
RQ and SQ follow the panoptic segmentation metrics of~\cite{kirillov2019panoptic}, while Boundary SQ denotes SQ computed under the boundary-aware IoU criterion using the Boundary IoU API~\cite{cheng2021boundary}. 
The results in Table~\ref{tab:decoder}-right show that designs b-1 and b-2 achieve moderately lower RQ than our design-a,
and significantly lower w.r.t the SQ and Boundary SQ metrics, confirming that our decoder design outputs masks with finer boundaries.
}

\begin{table}[t]
\centering
\vspace{-0mm}
\caption{\small
    \textbf{Finetuning pretrained image encoders is crucial for ICS.}
    \textbf{Left:} 
    Even powerful pretrained image encoders need adaptation or finetuning for effective ICS, while most existing approaches freeze pretrained image encoders (e.g., Matcher, SINE, SegIC, and INSID3).
    \textbf{Right:} For a given image, 
    we visualize the first three principal components of feature maps extracted by the pretrained and finetuned image encoders.
    Features maps produced by finetuned encoders exhibit salient groups of pixels w.r.t the object and its parts.
    }
\vspace{-3mm}
\begin{minipage}{0.43\textwidth}
\centering
\scalebox{0.70}{
    \setlength{\tabcolsep}{3.5pt}
    \renewcommand{\arraystretch}{1.1}

        \begin{tabular}{crr crr crr}
            \toprule
            \multicolumn{3}{c}{DINOv3}  &
            \multicolumn{3}{c}{PixIO}  &
            \multicolumn{3}{c}{$\text{PE}_{\text{spatial}}$}  \\
    \cmidrule(lr){1-3}
    \cmidrule(lr){4-6}
    \cmidrule(lr){7-9}
    
            ft? & DRAM & TigD  & 
            ft? & DRAM & TigD  & 
            ft? & DRAM & TigD  \\

            \ding{55} & 1.83 & 0.46 &
            \ding{55} & 14.45 & 7.31 &
            \ding{55} & 10.34 & 6.48
            \\ 
            
            \rowcolor{col1} $\checkmark$ & \textbf{59.20} & 12.69 &
            $\checkmark$ & 40.83 & 10.08 &
            $\checkmark$ & 45.02 & \textbf{13.83}
            \\
           
            \bottomrule
        \end{tabular}
    
}
\end{minipage}
\hfill
\begin{minipage}{0.55\textwidth}
    \raggedleft
    \includegraphics[width=0.95\linewidth]{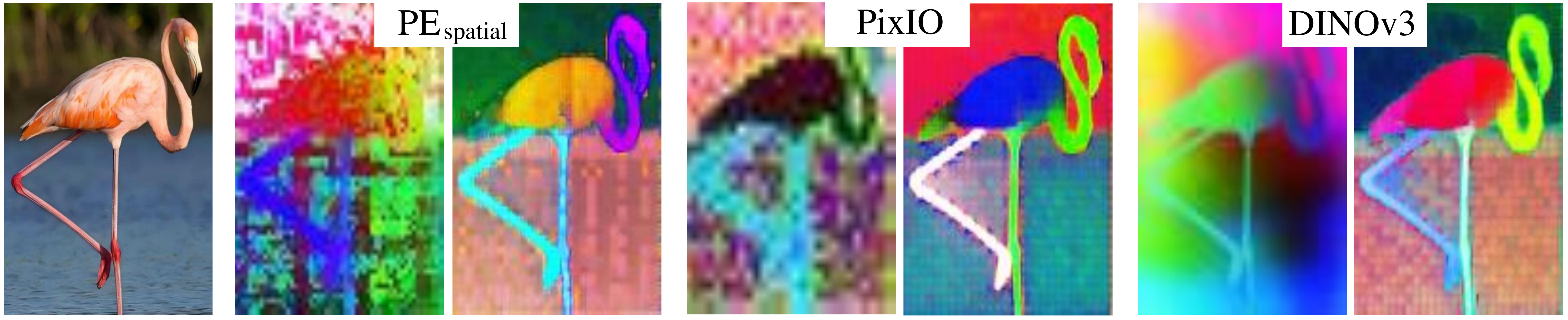}
\end{minipage}
\vspace{-3mm}
\label{tab:img-enc}
\end{table}

{\bf Finetuning the pretrained image encoder is critical for ICS.}
We justify this by initializing encoders from different pretrained ViTs, including
DINOv3,
PixIO~\cite{yang2026pursuit}, and 
$\text{PE}_{\text{spatial}}$~\cite{bolya2026perception},
and comparing frozen versus finetuned variants (Table~\ref{tab:img-enc}-left).
While most existing methods freeze the encoder \cite{balazevic2023towards, liu2023matcher, meng2024segic, liu2024simple, cuttano2026insid3}, 
finetuning consistently yields significant performance gains across encoders.
PCA visualizations of feature maps before and after finetuning (Table~\ref{tab:img-enc}-right) further reveal that finetuning sharpens the encoder's ability to delineate both holistic objects and their parts.
Together, these results confirm that
even powerful pretrained backbones need adaptation for effective ICS.

\begin{wrapfigure}{r}{0.4\textwidth}
{
\setlength{\tabcolsep}{1.15em} 
\centering
\vspace{-4mm}
\captionof{table}{\small temperature \& bias}
\label{tab:bias}
\vspace{-3.3mm}
\scalebox{0.80}{
    \renewcommand{\arraystretch}{1.1}
    \setlength{\tabcolsep}{3pt}
    \begin{tabular}{ccccccc}
        \toprule
         temp. & bias & DRAM$\uparrow$ & TigD$\uparrow$ & \revised{Ant$\uparrow$} & \revised{LDA$\uparrow$} & \revised{Gala$\uparrow$}\\
        \midrule
        \ding{55} & \ding{55} & fail & fail & fail & fail & fail\\
        $\checkmark$ & \ding{55} & 10.79 & \textcolor{white}{0}7.76 & \textcolor{white}{0}5.27 & 10.86 & \textcolor{white}{0}4.37\\
        \ding{55} & $\checkmark$ & fail & fail& fail & fail & fail\\
        \rowcolor{col1} $\checkmark$ & $\checkmark$ & \textbf{59.20} & \textbf{12.69} & \textbf{31.76} & \textbf{39.85} & \textbf{10.54}\\
        \bottomrule
    \end{tabular}
}
\vspace{-4mm}
}
\end{wrapfigure}
\textbf{Learning the temperature and bias is crucial.}
Table~\ref{tab:bias} shows that training fails 
without learning 
a temperature,
which strengthens supervision and facilitates training \cite{liu2026solving}.
Moreover, introducing a learnable bias boosts performance. 
We conjecture that, 
when classification is based on the cosine similarity between prototypes and mask query features,
a learnable bias properly calibrate the logit range to facilitate effective training.
Notably, DINOv omit this bias term \cite{DINOv_github_score, DINOv_github_sigmoid},
likely explaining its inferior performance compared to ours.

\begin{wrapfigure}{r}{0.36\textwidth}
{
\setlength{\tabcolsep}{0.65em} 
\centering
\vspace{-2mm}
\captionof{table}{\small  compute loss for \textit{ignota}?}
\label{tab:ignota}
\vspace{-3mm}
\scalebox{0.8}{
    \renewcommand{\arraystretch}{1.1}
    \setlength{\tabcolsep}{4pt}
        \begin{tabular}{cccccc}
        \toprule
         & DRAM$\uparrow$ & TigD$\uparrow$ & \revised{Ant$\uparrow$} & \revised{LDA$\uparrow$} & \revised{Gala$\uparrow$}\\
        \midrule
        \ding{55} & 54.25 & 10.72 & 27.60 & 34.06 & \textcolor{white}{0}3.24\\ 
        \rowcolor{col1} $\checkmark$ & \textbf{59.20} & \textbf{12.69} & \textbf{31.76} & \textbf{39.85} & \textbf{10.54}\\
        \bottomrule
    \end{tabular}
}
\vspace{-2mm}
}
\end{wrapfigure}
\textbf{Learning with \textit{ignota} helps train better \ourmodel.}
Recall that \textit{ignota} arise during training because randomly sampled query images often contain object classes not demonstrated by the randomly sampled prompt images.
For example, columns 1-3 of Fig.~\ref{fig:splashing_figure} show that zebras; grass and houses; and  
clock and lights are absent from their corresponding prompt images, thus becoming \textit{ignota}.
Under the BCE loss formulation,
\textit{ignota} can either be excluded from loss computation or treated as negative examples for all target classes.
Table~\ref{tab:ignota} shows that the latter approach facilitates the training of \ourmodel.

\begin{wrapfigure}{r}{0.45\textwidth}
{
\setlength{\tabcolsep}{0.05em} 
\centering
\vspace{-4mm}
\captionof{table}{\small classification loss.}
\label{tab:loss}
\vspace{-3.3mm}
\scalebox{0.8}{
    \renewcommand{\arraystretch}{1.1}
    \setlength{\tabcolsep}{4pt}
    \begin{tabular}{lcccccc}
        \toprule
         loss & \textit{ignota} & DRAM$\uparrow$ & TigD$\uparrow$ & \revised{Ant$\uparrow$} & \revised{LDA$\uparrow$} & \revised{Gala$\uparrow$}\\
        \midrule
        
        CE  & \ding{55} & 31.44 & \textcolor{white}{0}1.41 & 10.80 & 12.12 & \textcolor{white}{0}3.41\\
        
        CE  & $\checkmark$ & 37.59 & \textcolor{white}{0}2.20 & 11.92 & 19.28 & \textcolor{white}{0}2.72\\

       BCE & \ding{55} & 54.25 & 10.72 & 27.60 & 34.06 & \textcolor{white}{0}3.24\\
       
       \rowcolor{col1}  BCE & $\checkmark$ & \textbf{59.20} & \textbf{12.69} & \textbf{31.76} & \textbf{39.85} & \textbf{10.54}\\
        \bottomrule
    \end{tabular}
}
\vspace{-2mm}
}
\end{wrapfigure}
\textbf{BCE loss vs. CE loss.}
Cross-entropy (CE) loss is widely used in recent influential methods, 
including MaskFormer and Mask2Former for general image segmentation,
CLIP for learning visual and text encoders from image-text pairs,
X-Decoder for foundational vision and vision-language tasks,
and the ICS method SINE  \cite{liu2024simple}.
However, 
Table~\ref{tab:loss} demonstrates that BCE loss is more appropriate than CE loss for training \ourmodel.
\revised{Notably, we evaluate CE and BCE loss under two different executions for the \textit{ignota} regions:
either ignoring them entirely in loss computation or modeling them as a separate placeholder class.
As shown in the results, BCE loss with explicit ignota supervision performs the best.}

\begin{wrapfigure}{r}{0.35\textwidth}
\centering
\vspace{-5mm}
\includegraphics[width=0.99\linewidth]{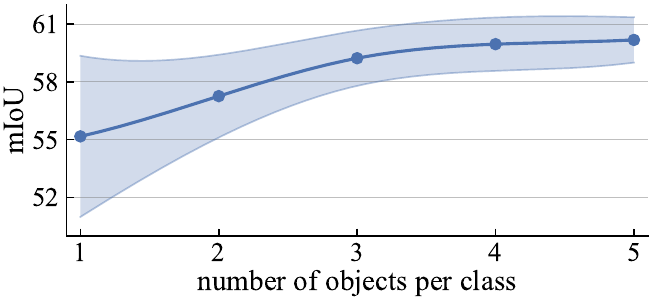}
\vspace{-6.5mm}
\caption{\small
\ourmodel performs better with more prompts.
}
\vspace{-10mm}
\label{fig:few-shot} 
\end{wrapfigure}
\textbf{\ourmodel's performance vs. visual demo size.}
Thanks to the lightweight mask decoder,
\ourmodel can process a large number of prompt images with negligible computational overhead.
We evaluate this on the 12-class DRAM dataset by varying the number of prompts.
Specifically,
from DRAM's training set, we randomly sample prompt images and  ensure that they contain exactly $n$ examples for each class.
We vary $n$ from 1 to 5.
For each $n$, we sample 5 random sets of prompts from the training set and report the averaged mIoU over the 5 sets on DRAM's test set.
Fig.~\ref{fig:few-shot} plots the averaged mIoU with standard deviation as a function of $n$,
showing that increasing the number of prompts improves  \ourmodel's performance.

\section{Conclusions}

We presented SAYD, a foundational in-context segmentation model that learns from heterogeneous prompt-query pairs.
By developing specialized architectural and training strategies, 
SAYD bypasses the limitations of language-dependent models to capture tacit visual knowledge that is readily recognized by human experts but impossible to articulate in text.
Extensive experiments across 24 benchmarks spanning natural, medical, ecological, and remote sensing domains demonstrate that SAYD significantly outperforms existing in-context, open-vocabulary, and SAM-based image segmentation methods. 
SAYD establishes a robust paradigm for in-context visual learning, enabling segmenting the nuanced, unspoken visual patterns critical to specialized expert domains.

\subsection*{AI use statement}

Generative AI tools were utilized strictly for language refinement and grammar corrections.
The authors thoroughly reviewed all AI-assisted revisions to ensure they precisely reflected the original concepts, methodology descriptions, and conclusions of this work.
AI tools were not used to generate research hypotheses, data, experimental designs and results, figures, tables, code, or any core technical contributions.

\subsection*{Ethics statement}
This work does not involve human subjects or the collection of new datasets.
Our method builds upon the open-source
vision foundation models and vision-language models,
and public datasets under their original licenses. 
Our method does not attempt to infer demographics, identities, or other sensitive attributes.

\subsection*{Reproducibility statement}

We promote reproducible research.
In this paper, we disclose important details of training in  \S\ref{ssec:implementation} along with more details in Appendix \S\ref{sec_suppl:arch} and 
dataset details in Appendix \S\ref{sec_suppl:datasets}.
Importantly,
we provide our code (described in Appendix \S\ref{sec_suppl:code}),
including Jupyter Notebook files that demonstrate not only code but also intermediate results.
We will release our open-source code and models to the public under the MIT License.

\subsection*{Acknowledgments}
The authors thank anonymous reviewers for their constructive feedback, which substantially improved the manuscript. 
SK gratefully acknowledges support from 
the CK Foundation and the Plato Initiative.
This work was supported by 
the Science and Technology Development Fund of Macau (0058/2025/RIA2, 0067/2024/ITP2),  and
the University of Macau (MYRG-CRG2026-00038-FIC, MYRG-GRG2026-00236-FIC, SRG2023-00044-FST).



\bibliography{arxiv}

@article{lenat1985cyc,
  title={CYC: Using common sense knowledge to overcome brittleness and knowledge acquisition bottlenecks},
  author={Lenat, Douglas B and Prakash, Mayank and Shepherd, Mary},
  journal={AI magazine},
  volume={6},
  number={4},
  pages={65--65},
  year={1985}
}

@article{meltzoff1999imitation,
  title={Imitation in infancy},
  author={Meltzoff, A and Moore, M and Nadel, J and Butterworth, G},
  journal={Psychology: Revisiting the Classic Studies},
  pages={71},
  year={1999}
}

@article{lavi2018developmental,
  title={Developmental differences between children and adults in the use of visual cues for segmentation},
  author={Lavi-Rotbain, Ori and Arnon, Inbal},
  journal={Cognitive science},
  volume={42},
  pages={606--620},
  year={2018},
  publisher={Wiley Online Library}
}

@article{johnson2010infants,
  title={How infants learn about the visual world},
  author={Johnson, Scott P},
  journal={Cognitive science},
  volume={34},
  number={7},
  pages={1158--1184},
  year={2010},
  publisher={Wiley Online Library}
}

@inproceedings{hanney1997adaptation,
  title={The adaptation knowledge bottleneck: How to ease it by learning from cases},
  author={Hanney, Kathleen and Keane, Mark T},
  booktitle={International Conference on Case-Based Reasoning},
  pages={359--370},
  year={1997},
  organization={Springer}
}

@article{katzke2026high,
  title={High-throughput phenomics of global ant biodiversity},
  author={Katzke, Julian and Hita Garcia, Francisco and L{\"o}sel, Philipp D and Azuma, Fumika and Farag{\'o}, Tom{\'a}{\v{s}} and Aibekova, Lazzat and Casadei-Ferreira, Alexandre and Gautam, Shubham and Richter, Adrian and Toulkeridou, Evropi and others},
  journal={Nature methods},
  volume={23},
  number={3},
  pages={663--672},
  year={2026}
}

@article{collins1974tea,
  title={The TEA set: Tacit knowledge and scientific networks},
  author={Collins, Harry M},
  journal={Science studies},
  volume={4},
  number={2},
  pages={165--185},
  year={1974},
  publisher={Sage Publications Sage CA: Thousand Oaks, CA}
}

@incollection{patel1999expertise,
  title={Expertise and tacit knowledge in medicine},
  author={Patel, Vimla L and Arocha, Jose F and Kaufman, David R},
  booktitle={Tacit knowledge in professional practice},
  pages={89--114},
  year={1999},
  publisher={Psychology Press}
}

@article{thornton2006tacit,
  title={Tacit knowledge as the unifying factor in evidence based medicine and clinical judgement},
  author={Thornton, Tim},
  journal={Philosophy, Ethics, and Humanities in Medicine},
  volume={1},
  number={1},
  pages={2},
  year={2006},
  publisher={Springer}
}

@article{heiberg2008tacit,
  title={Tacit knowledge and visual expertise in medical diagnostic reasoning: implications for medical education},
  author={Heiberg Engel, Peter Johan},
  journal={Medical teacher},
  volume={30},
  number={7},
  pages={e184--e188},
  year={2008},
  publisher={Taylor \& Francis}
}

@inproceedings{kong2018recurrent,
  title={Recurrent pixel embedding for instance grouping},
  author={Kong, Shu and Fowlkes, Charless C},
  booktitle={Proceedings of the IEEE conference on computer vision and pattern recognition},
  year={2018}
}

@inproceedings{kirillov2023segment,
  title={Segment anything},
  author={Kirillov, Alexander and Mintun, Eric and Ravi, Nikhila and Mao, Hanzi and Rolland, Chloe and Gustafson, Laura and Xiao, Tete and Whitehead, Spencer and Berg, Alexander C and Lo, Wan-Yen and others},
  booktitle={Proceedings of the IEEE/CVF international conference on computer vision},
  pages={4015--4026},
  year={2023}
}

@article{simeoni2025dinov3,
  title={Dinov3},
  author={Oriane, Sim{\'e}oni and Vo, Huy V and Seitzer, Maximilian and Baldassarre, Federico and Oquab, Maxime and Jose, Cijo and Khalidov, Vasil and Szafraniec, Marc and Yi, Seungeun and Ramamonjisoa, Micha{\"e}l and others},
  journal={arXiv preprint arXiv:2508.10104},
  year={2025}
}

@article{ren2024grounded,
      title={Grounded SAM: Assembling Open-World Models for Diverse Visual Tasks}, 
      author={Tianhe Ren and Shilong Liu and Ailing Zeng and Jing Lin and Kunchang Li and He Cao and Jiayu Chen and Xinyu Huang and Yukang Chen and Feng Yan and Zhaoyang Zeng and Hao Zhang and Feng Li and Jie Yang and Hongyang Li and Qing Jiang and Lei Zhang},
      journal={arXiv:2401.14159},
      year={2024}
}

@inproceedings{zou2023generalized,
  title={Generalized decoding for pixel, image, and language},
  author={Zou, Xueyan and Dou, Zi-Yi and Yang, Jianwei and Gan, Zhe and Li, Linjie and Li, Chunyuan and Dai, Xiyang and Behl, Harkirat and Wang, Jianfeng and Yuan, Lu and others},
  booktitle={Proceedings of the IEEE/CVF conference on computer vision and pattern recognition},
  pages={15116--15127},
  year={2023}
}

@inproceedings{shen2024aligning,
  title={Aligning and prompting everything all at once for universal visual perception},
  author={Shen, Yunhang and Fu, Chaoyou and Chen, Peixian and Zhang, Mengdan and Li, Ke and Sun, Xing and Wu, Yunsheng and Lin, Shaohui and Ji, Rongrong},
  booktitle={Proceedings of the IEEE/CVF Conference on Computer Vision and Pattern Recognition},
  pages={13193--13203},
  year={2024}
}

@inproceedings{zhang2023simple,
  title={A simple framework for open-vocabulary segmentation and detection},
  author={Hao, Zhang and Li, Feng and Zou, Xueyan and Liu, Shilong and Li, Chunyuan and Yang, Jianwei and Zhang, Lei},
  booktitle={Proceedings of the IEEE/CVF International Conference on Computer Vision},
  pages={1020--1031},
  year={2023}
}

@inproceedings{xiao2025openworldsam,
  title     = {OpenWorldSAM: Extending SAM2 for Universal Image Segmentation with Language Prompts},
  author={Xiao, Shiting and Kabra, Rishabh and Li, Yuhang and Lee, Donghyun and Carreira, Joao and Panda, Priyadarshini},
  booktitle = {Advances in Neural Information Processing Systems (NeurIPS)},
  year      = {2025},
}

@inproceedings{alama2026radseg,
  title={RADSeg: Unleashing Parameter and Compute Efficient Zero-Shot Open-Vocabulary Segmentation Using Agglomerative Models},
  author={Omar, Alama and Jariwala, Darshil and Bhattacharya, Avigyan and Kim, Seungchan and Wang, Wenshan and Scherer, Sebastian},
  booktitle={Proceedings of the IEEE/CVF Conference on Computer Vision and Pattern Recognition},
  pages={9294--9304},
  year={2026}
}

@inproceedings{li2024visual,
  title={Visual in-context prompting},
  author={Li, Feng and Jiang, Qing and Zhang, Hao and Ren, Tianhe and Liu, Shilong and Zou, Xueyan and Xu, Huaizhe and Li, Hongyang and Yang, Jianwei and Li, Chunyuan and others},
  booktitle={Proceedings of the IEEE/CVF Conference on Computer Vision and Pattern Recognition},
  pages={12861--12871},
  year={2024}
}

@inproceedings{wang2023images,
  title={Images speak in images: A generalist painter for in-context visual learning},
  author={Wang, Xinlong and Wang, Wen and Cao, Yue and Shen, Chunhua and Huang, Tiejun},
  booktitle={Proceedings of the IEEE/CVF Conference on Computer Vision and Pattern Recognition},
  pages={6830--6839},
  year={2023}
}

@article{wang2023seggpt,
  title={Seggpt: Segmenting everything in context},
  author={Wang, Xinlong and Zhang, Xiaosong and Cao, Yue and Wang, Wen and Shen, Chunhua and Huang, Tiejun},
  journal={arXiv preprint arXiv:2304.03284},
  year={2023}
}

@inproceedings{carion2025sam,
  title={Sam 3: Segment anything with concepts},
  author={Carion, Nicolas and Gustafson, Laura and Hu, Yuan-Ting and Debnath, Shoubhik and Hu, Ronghang and Suris, Didac and Ryali, Chaitanya and Alwala, Kalyan Vasudev and Khedr, Haitham and Huang, Andrew and others},
  booktitle={International Conference on Learning Representations (ICLR)},
  year={2026}
}

@inproceedings{he2022masked,
  title={Masked autoencoders are scalable vision learners},
  author={He, Kaiming and Chen, Xinlei and Xie, Saining and Li, Yanghao and Doll{\'a}r, Piotr and Girshick, Ross},
  booktitle={Proceedings of the IEEE/CVF Conference on Computer Vision and Pattern Recognition (CVPR)},
  pages={16000--16009},
  year={2022}
}

@inproceedings{yang2026pursuit,
  title={In pursuit of pixel supervision for visual pre-training},
  author={Yang, Lihe and Li, Shang-Wen and Li, Yang and Lei, Xinjie and Wang, Dong and Mohamed, Abdelrahman and Xie, Saining and Zhao, Hengshuang and He, Kaiming and Xu, Hu},
  booktitle={Proceedings of the IEEE/CVF Conference on Computer Vision and Pattern Recognition},
  pages={31974--31984},
  year={2026}
}

@InProceedings{chen2021mocov3,
  author  = {Xinlei Chen and Saining Xie and Kaiming He},
  title   = {An Empirical Study of Training Self-Supervised Vision Transformers},
  booktitle = {International Conference on Computer Vision (ICCV)},
  year    = {2021},
}

@inproceedings{he2020momentum,
  title={Momentum contrast for unsupervised visual representation learning},
  author={He, Kaiming and Fan, Haoqi and Wu, Yuxin and Xie, Saining and Girshick, Ross},
  booktitle={Proceedings of the IEEE/CVF Conference on Computer Vision and Pattern Recognition (CVPR)},
  pages={9729--9738},
  year={2020}
}

@article{oquab2023dinov2,
  title={Dinov2: Learning robust visual features without supervision},
  author={Oquab, Maxime and Darcet, Timoth{\'e}e and Moutakanni, Th{\'e}o and Vo, Huy and Szafraniec, Marc and Khalidov, Vasil and Fernandez, Pierre and Haziza, Daniel and Massa, Francisco and El-Nouby, Alaaeldin and others},
  journal={arXiv preprint arXiv:2304.07193},
  year={2023}
}

@article{brown2020language,
  title={Language models are few-shot learners},
  author={Brown, Tom and Mann, Benjamin and Ryder, Nick and Subbiah, Melanie and Kaplan, Jared D and Dhariwal, Prafulla and Neelakantan, Arvind and Shyam, Pranav and Sastry, Girish and Askell, Amanda and others},
  journal={Advances in neural information processing systems},
  volume={33},
  pages={1877--1901},
  year={2020}
}

@inproceedings{radford2021learning,
  title={Learning transferable visual models from natural language supervision},
  author={Radford, Alec and Kim, Jong Wook and Hallacy, Chris and Ramesh, Aditya and Goh, Gabriel and Agarwal, Sandhini and Sastry, Girish and Askell, Amanda and Mishkin, Pamela and Clark, Jack and others},
  booktitle={International conference on machine learning},
  pages={8748--8763},
  year={2021},
  organization={PmLR}
}

@inproceedings{xu2024demystifying,
  title={Demystifying clip data},
  author={Xu, Hu and Xie, Saining and Tan, Xiaoqing and Huang, Po-Yao and Howes, Russell and Sharma, Vasu and Li, Shang-Wen and Ghosh, Gargi and Zettlemoyer, Luke and Feichtenhofer, Christoph},
  booktitle={International Conference on Learning Representations},
  volume={2024},
  pages={47812--47831},
  year={2024}
}

@article{chuang2026meta,
  title={Meta clip 2: A worldwide scaling recipe},
  author={Chuang, Yung-Sung and Li, Yang and Wang, Dong and Yeh, Ching-Feng and Lyu, Kehan and Raghavendra, Ramya and Glass, Jim and Huang, Lifei and Weston, Jason and Zettlemoyer, Luke and others},
  journal={Advances in Neural Information Processing Systems},
  volume={38},
  pages={48009--48036},
  year={2026}
}

@inproceedings{liu2026solving,
title={Solving Semi-Supervised Few-Shot Learning from an Auto-Annotation Perspective},
author={Liu, Tian and Basu, Anwesha and Caverlee, James and Kong, Shu},
booktitle={European Conference on Computer Vision (ECCV)},
year={2026}
}

@misc{DINOv_github_score,
  author={DINOv},
  title = {GitHub},
  howpublished = {\url{{https://github.com/UX-Decoder/DINOv/blob/main/dinov/body/decoder/dinov_openset_decoder.py#L966}}},
  note = {commit: 49fd351},
  year={2024}
}

@misc{DINOv_github_sigmoid,
  author={DINOv},
  title = {GitHub},
  howpublished = {\url{{https://github.com/UX-Decoder/DINOv/blob/main/dinov/modules/criterion_visual_openset.py#L126}}},
  note = {commit: 49fd351},
  year={2024}
}

@article{scheirer2012toward,
  title={Toward open set recognition},
  author={Scheirer, Walter J and de Rezende Rocha, Anderson and Sapkota, Archana and Boult, Terrance E},
  journal={IEEE transactions on pattern analysis and machine intelligence},
  volume={35},
  number={7},
  pages={1757--1772},
  year={2012}
}

@misc{qwen3.8,
    title={{Qwen3.8-Max}: A New Bar for Coding and Cowork},
    author={{Qwen Team}},
    year={2026},
    month={August},
    url={https://qwen.ai/blog?id=qwen3.8}
}

@article{adaime2024deep,
  title={Deep learning approaches to the phylogenetic placement of extinct pollen morphotypes},
  author={Ada{\"\i}m{\'e}, Marc-{\'E}lie and Kong, Shu and Punyasena, Surangi W},
  journal={PNAS nexus},
  volume={3},
  number={1},
  pages={pgad419},
  year={2024},
  publisher={Oxford University Press US}
}

@article{chakravarthy2025lidar,
  title={Lidar panoptic segmentation in an open world},
  author={Chakravarthy, Anirudh S and Ganesina, Meghana Reddy and Hu, Peiyun and Leal-Taix{\'e}, Laura and Kong, Shu and Ramanan, Deva and Osep, Aljosa},
  journal={International Journal of Computer Vision},
  volume={133},
  number={3},
  pages={1153--1174},
  year={2025},
  publisher={Springer}
}

@article{wei2023ov,
  title={Ov-parts: Towards open-vocabulary part segmentation},
  author={Wei, Meng and Yue, Xiaoyu and Zhang, Wenwei and Kong, Shu and Liu, Xihui and Pang, Jiangmiao},
  journal={Advances in Neural Information Processing Systems},
  volume={36},
  pages={70094--70114},
  year={2023}
}

@inproceedings{ma2026autoexpert,
  title     = {{AutoExpert}: Automating 3D {LiDAR} Annotation from Expert-Crafted Guidelines},
  author    = {Ma, Yechi and Hua, Wei and Kong, Shu},
  booktitle = {Advances in Neural Information Processing Systems},
  year      = {2026}}

@inproceedings{shi2024lca,
  title={Lca-on-the-line: Benchmarking out-of-distribution generalization with class taxonomies},
  author={Shi, Jia and Gare, Gautam and Tian, Jinjin and Chai, Siqi and Lin, Zhiqiu and Vasudevan, Arun and Feng, Di and Ferroni, Francesco and Kong, Shu},
  booktitle={Proceedings of the 41st International Conference on Machine Learning},
  year={2024}
}

@inproceedings{kong2021opengan,
  title={Opengan: Open-set recognition via open data generation},
  author={Kong, Shu and Ramanan, Deva},
  booktitle={Proceedings of the IEEE/CVF international conference on computer vision},
  year={2021}
}

@misc{SINE_github_sampling,
  author={SINE},
  title = {GitHub},
  howpublished = {\url{{https://github.com/aim-uofa/SINE/blob/main/sine/model/model.py#L118}}},
  note = {commit: a7a989c},
  year={2024}
}

@article{biagioli1995tacit,
  title={Tacit knowledge, courtliness, and the scientist’s body},
  author={Biagioli, Mario},
  journal={Choreographing history},
  pages={69--81},
  year={1995},
  publisher={Indiana University Press Bloomington and Indianapolis}
}

@article{bolya2026perception,
  title={Perception encoder: The best visual embeddings are not at the output of the network},
  author={Bolya, Daniel and Huang, Po-Yao and Sun, Peize and Cho, Jang Hyun and Madotto, Andrea and Wei, Chen and Ma, Tengyu and Zhi, Jiale and Rajasegaran, Jathushan and Bangalath, Hanoona and others},
  journal={Advances in Neural Information Processing Systems},
  volume={38},
  pages={60884--60937},
  year={2026}
}

@article{schuhmann2022laion,
  title={Laion-5b: An open large-scale dataset for training next generation image-text models},
  author={Schuhmann, Christoph and Beaumont, Romain and Vencu, Richard and Gordon, Cade and Wightman, Ross and Cherti, Mehdi and Coombes, Theo and Katta, Aarush and Mullis, Clayton and Wortsman, Mitchell and others},
  journal={Advances in neural information processing systems},
  volume={35},
  pages={25278--25294},
  year={2022}
}

@article{bar2022visual,
  title={Visual prompting via image inpainting},
  author={Bar, Amir and Gandelsman, Yossi and Darrell, Trevor and Globerson, Amir and Efros, Alexei},
  journal={Advances in neural information processing systems},
  volume={35},
  pages={25005--25017},
  year={2022}
}

@inproceedings{carion2020end,
  title={End-to-end object detection with transformers},
  author={Carion, Nicolas and Massa, Francisco and Synnaeve, Gabriel and Usunier, Nicolas and Kirillov, Alexander and Zagoruyko, Sergey},
  booktitle={European conference on computer vision},
  pages={213--229},
  year={2020},
  organization={Springer}
}

@inproceedings{caron2021emerging,
  title={Emerging properties in self-supervised vision transformers},
  author={Caron, Mathilde and Touvron, Hugo and Misra, Ishan and J{\'e}gou, Herv{\'e} and Mairal, Julien and Bojanowski, Piotr and Joulin, Armand},
  booktitle={Proceedings of the IEEE/CVF international conference on computer vision},
  pages={9650--9660},
  year={2021}
}

@inproceedings{zhou2017scene,
  title={Scene parsing through ade20k dataset},
  author={Zhou, Bolei and Zhao, Hang and Puig, Xavier and Fidler, Sanja and Barriuso, Adela and Torralba, Antonio},
  booktitle={Proceedings of the IEEE conference on computer vision and pattern recognition},
  pages={633--641},
  year={2017}
}

@inproceedings{lin2014microsoft,
  title={Microsoft coco: Common objects in context},
  author={Lin, Tsung-Yi and Maire, Michael and Belongie, Serge and Hays, James and Perona, Pietro and Ramanan, Deva and Doll{\'a}r, Piotr and Zitnick, C Lawrence},
  booktitle={European conference on computer vision},
  pages={740--755},
  year={2014},
  organization={Springer}
}

@inproceedings{zhai2023sigmoid,
  title={Sigmoid loss for language image pre-training},
  author={Zhai, Xiaohua and Mustafa, Basil and Kolesnikov, Alexander and Beyer, Lucas},
  booktitle={Proceedings of the IEEE/CVF international conference on computer vision},
  pages={11975--11986},
  year={2023}
}

@article{cheng2021per,
  title={Per-pixel classification is not all you need for semantic segmentation},
  author={Cheng, Bowen and Schwing, Alex and Kirillov, Alexander},
  journal={Advances in neural information processing systems},
  volume={34},
  pages={17864--17875},
  year={2021}
}

@inproceedings{cuttano2026insid3,
  title={INSID3: Training-Free In-Context Segmentation with DINOv3},
  author={Cua, Claudia and Trivigno, Gabriele and Reich, Christoph and Cremers, Daniel and Masone, Carlo and Roth, Stefan},
  booktitle={Proceedings of the IEEE/CVF Conference on Computer Vision and Pattern Recognition},
  pages={21638--21648},
  year={2026}
}

@article{liu2024simple,
  title={A simple image segmentation framework via in-context examples},
  author={Liu, Yang and Jing, Chenchen and Li, Hengtao and Zhu, Muzhi and Chen, Hao and Wang, Xinlong and Shen, Chunhua},
  journal={Advances in Neural Information Processing Systems},
  volume={37},
  pages={25095--25119},
  year={2024}
}

@article{liu2023matcher,
  title={Matcher: Segment anything with one shot using all-purpose feature matching},
  author={Liu, Yang and Zhu, Muzhi and Li, Hengtao and Chen, Hao and Wang, Xinlong and Shen, Chunhua},
  journal={arXiv preprint arXiv:2305.13310},
  year={2023}
}

@inproceedings{meng2024segic,
  title={Segic: Unleashing the emergent correspondence for in-context segmentation},
  author={Meng, Lingchen and Lan, Shiyi and Li, Hengduo and Alvarez, Jose M and Wu, Zuxuan and Jiang, Yu-Gang},
  booktitle={European Conference on Computer Vision},
  pages={203--220},
  year={2024},
  organization={Springer}
}

@article{balazevic2023towards,
  title={Towards in-context scene understanding},
  author={Balazevic, Ivana and Steiner, David and Parthasarathy, Nikhil and Arandjelovi{\'c}, Relja and Henaff, Olivier},
  journal={Advances in Neural Information Processing Systems},
  volume={36},
  pages={63758--63778},
  year={2023}
}

@article{dosovitskiy2020image,
  title={An image is worth 16x16 words: Transformers for image recognition at scale},
  author={Dosovitskiy, Alexey and Beyer, Lucas and Kolesnikov, Alexander and Weissenborn, Dirk and Zhai, Xiaohua and Unterthiner, Thomas and Dehghani, Mostafa and Minderer, Matthias and Heigold, Georg and Gelly, Sylvain and others},
  journal={arXiv preprint arXiv:2010.11929},
  year={2020}
}

@inproceedings{li2022exploring,
  title={Exploring plain vision transformer backbones for object detection},
  author={Li, Yanghao and Mao, Hanzi and Girshick, Ross and He, Kaiming},
  booktitle={European conference on computer vision},
  pages={280--296},
  year={2022},
  organization={Springer}
}

@inproceedings{cheng2022masked,
  title={Masked-attention mask transformer for universal image segmentation},
  author={Cheng, Bowen and Misra, Ishan and Schwing, Alexander G and Kirillov, Alexander and Girdhar, Rohit},
  booktitle={Proceedings of the IEEE/CVF conference on computer vision and pattern recognition},
  pages={1290--1299},
  year={2022}
}

@inproceedings{wu2018unsupervised,
  title={Unsupervised feature learning via non-parametric instance discrimination},
  author={Wu, Zhirong and Xiong, Yuanjun and Yu, Stella X and Lin, Dahua},
  booktitle={Proceedings of the IEEE conference on computer vision and pattern recognition},
  pages={3733--3742},
  year={2018}
}

@inproceedings{lin2017focal,
  title={Focal loss for dense object detection},
  author={Lin, Tsung-Yi and Goyal, Priya and Girshick, Ross and He, Kaiming and Doll{\'a}r, Piotr},
  booktitle={Proceedings of the IEEE international conference on computer vision},
  pages={2980--2988},
  year={2017}
}

@inproceedings{Cordts2016Cityscapes,
title={The Cityscapes Dataset for Semantic Urban Scene Understanding},
author={Cordts, Marius and Omran, Mohamed and Ramos, Sebastian and Rehfeld, Timo and Enzweiler, Markus and Benenson, Rodrigo and Franke, Uwe and Roth, Stefan and Schiele, Bernt},
booktitle={Proc. of the IEEE Conference on Computer Vision and Pattern Recognition (CVPR)},
year={2016}
}

@inproceedings{richter2016playing,
  title={Playing for data: Ground truth from computer games},
  author={Richter, Stephan R and Vineet, Vibhav and Roth, Stefan and Koltun, Vladlen},
  booktitle={European conference on computer vision},
  pages={102--118},
  year={2016},
  organization={Springer}
}

@inproceedings{cohen2022semantic,
  title={Semantic segmentation in art paintings},
  author={Cohen, Nadav and Newman, Yael and Shamir, Ariel},
  booktitle={Computer graphics forum},
  volume={41},
  pages={261--275},
  year={2022},
  organization={Wiley Online Library}
}

@article{wah2011caltech,
  title={The caltech-ucsd birds-200-2011 dataset},
  author={Wah, Catherine and Branson, Steve and Welinder, Peter and Perona, Pietro and Belongie, Serge},
  year={2011}
}

@article{wei2409plantseg,
  title={Plantseg: A large-scale in-the-wild dataset for plant disease segmentation. arXiv 2024},
  author={Wei, T and Chen, Z and Yu, X and Chapman, S and Melloy, P and Huang, Z},
  journal={arXiv preprint arXiv:2409.04038},
  year={2024}
}

@inproceedings{gong2017look,
  title={Look into person: Self-supervised structure-sensitive learning and a new benchmark for human parsing},
  author={Gong, Ke and Liang, Xiaodan and Zhang, Dongyu and Shen, Xiaohui and Lin, Liang},
  booktitle={Proceedings of the IEEE conference on computer vision and pattern recognition},
  pages={932--940},
  year={2017}
}

@inproceedings{mu2020learning,
  title={Learning from synthetic animals},
  author={Mu, Jiteng and Qiu, Weichao and Hager, Gregory D and Yuille, Alan L},
  booktitle={Proceedings of the IEEE/CVF Conference on Computer Vision and Pattern Recognition},
  pages={12386--12395},
  year={2020}
}

@article{loshchilov2017decoupled,
  title={Decoupled weight decay regularization},
  author={Loshchilov, Ilya and Hutter, Frank},
  journal={arXiv preprint arXiv:1711.05101},
  year={2017}
}

@article{goyal2017accurate,
  title={Accurate, large minibatch sgd: Training imagenet in 1 hour},
  author={Goyal, Priya and Doll{\'a}r, Piotr and Girshick, Ross and Noordhuis, Pieter and Wesolowski, Lukasz and Kyrola, Aapo and Tulloch, Andrew and Jia, Yangqing and He, Kaiming},
  journal={arXiv preprint arXiv:1706.02677},
  year={2017}
}

@article{bao2021beit,
  title={Beit: Bert pre-training of image transformers},
  author={Bao, Hangbo and Dong, Li and Piao, Songhao and Wei, Furu},
  journal={arXiv preprint arXiv:2106.08254},
  year={2021}
}

@inproceedings{Wang2021LoveDA,
    author = {Wang, Junjue and Zheng, Zhuo and Ma, Ailong and Lu, Xiaoyan and Zhong, Yanfei},
    booktitle = {Proceedings of the Neural Information Processing Systems Track on Datasets and Benchmarks},
    title = {LoveDA: A Remote Sensing Land-Cover Dataset for Domain Adaptive Semantic Segmentation},
    year = {2021}
}

@article{shiraishi2000development,
  title={Development of a digital image database for chest radiographs with and without a lung nodule: receiver operating characteristic analysis of radiologists' detection of pulmonary nodules},
  author={Shiraishi, Junji and Katsuragawa, Shigehiko and Ikezoe, Junpei and Matsumoto, Tsuneo and Kobayashi, Takeshi and Komatsu, Ken-ichi and Matsui, Mitate and Fujita, Hiroshi and Kodera, Yoshie and Doi, Kunio},
  journal={American journal of roentgenology},
  volume={174},
  number={1},
  pages={71--74},
  year={2000},
  publisher={American Roentgen Ray Society}
}

@misc{metaldam2024,
  title={MetalDAM: Metallography Dataset from Additive Manufacturing},
  author={ArcelorMittal},
  year={2024},
}

@article{edlund2021livecell,
  title={LIVECell—A large-scale dataset for label-free live cell segmentation},
  author={Edlund, Christoffer and Jackson, Timothy R and Khalid, Nabeel and Bevan, Nicola and Dale, Timothy and Dengel, Andreas and Ahmed, Sheraz and Trygg, Johan and Sj{\"o}gren, Rickard},
  journal={Nature methods},
  volume={18},
  number={9},
  pages={1038--1045},
  year={2021},
  publisher={Nature Publishing Group US New York}
}

@article{willett2013galaxy,
  title={Galaxy Zoo 2: detailed morphological classifications for 304 122 galaxies from the Sloan Digital Sky Survey},
  author={Willett, Kyle W and Lintott, Chris J and Bamford, Steven P and Masters, Karen L and Simmons, Brooke D and Casteels, Kevin RV and Edmondson, Edward M and Fortson, Lucy F and Kaviraj, Sugata and Keel, William C and others},
  journal={Monthly Notices of the Royal Astronomical Society},
  volume={435},
  number={4},
  pages={2835--2860},
  year={2013},
  publisher={Oxford University Press}
}

@inproceedings{liu2021swin,
  title={Swin transformer: Hierarchical vision transformer using shifted windows},
  author={Liu, Ze and Lin, Yutong and Cao, Yue and Hu, Han and Wei, Yixuan and Zhang, Zheng and Lin, Stephen and Guo, Baining},
  booktitle={Proceedings of the IEEE/CVF international conference on computer vision},
  pages={10012--10022},
  year={2021}
}

@inproceedings{deng2024coconut,
  title={Coconut: Modernizing coco segmentation},
  author={Deng, Xueqing and Yu, Qihang and Wang, Peng and Shen, Xiaohui and Chen, Liang-Chieh},
  booktitle={2024 IEEE/CVF Conference on Computer Vision and Pattern Recognition (CVPR)},
  pages={21863--21873},
  year={2024},
  organization={IEEE}
}

@inproceedings{neuhold2017mapillary,
  title={The mapillary vistas dataset for semantic understanding of street scenes},
  author={Neuhold, Gerhard and Ollmann, Tobias and Rota Bulo, Samuel and Kontschieder, Peter},
  booktitle={Proceedings of the IEEE international conference on computer vision},
  pages={4990--4999},
  year={2017}
}

@inproceedings{miao2022large,
  title={Large-scale video panoptic segmentation in the wild: A benchmark},
  author={Miao, Jiaxu and Wang, Xiaohan and Wu, Yu and Li, Wei and Zhang, Xu and Wei, Yunchao and Yang, Yi},
  booktitle={2022 IEEE/CVF Conference on Computer Vision and Pattern Recognition (CVPR)},
  pages={21001--21011},
  year={2022},
  organization={IEEE}
}

@inproceedings{le2024jrdbpanotrack,
    title={JRDB-PanoTrack: An Open-world Panoptic Segmentation and Tracking Robotic Dataset in Crowded Human Environments},
    author={Le, Duy-Tho and Gou, Chenhui and Datta, Stavya and Shi, Hengcan and Reid, Ian and Cai, Jianfei and Rezatofighi, Hamid},
    booktitle={Proceedings of the IEEE/CVF Conference on Computer Vision and Pattern Recognition},
    year={2024}
}

@InProceedings{Zust2023LaRS,
  title={LaRS: A Diverse Panoptic Maritime Obstacle Detection Dataset and Benchmark},
  author={{\v{Z}}ust, Lojze and Per{\v{s}}, Janez and Kristan, Matej},
  booktitle={International Conference on Computer Vision (ICCV)},
  year={2023}
}

@inproceedings{gong2018instance,
  title={Instance-level human parsing via part grouping network},
  author={Gong, Ke and Liang, Xiaodan and Li, Yicheng and Chen, Yimin and Yang, Ming and Lin, Liang},
  booktitle={European conference on computer vision},
  pages={805--822},
  year={2018},
  organization={Springer}
}

@article{he2021partimagenet,
  title={PartImageNet: A Large, High-Quality Dataset of Parts},
  author={He, Ju and Yang, Shuo and Yang, Shaokang and Kortylewski, Adam and Yuan, Xiaoding and Chen, Jie-Neng and Liu, Shuai and Yang, Cheng and Yuille, Alan},
  journal={arXiv preprint arXiv:2112.00933},
  year={2021}
}

@inproceedings{li2024pinpp,
  author = {Li, Xiao and Liu, Yining and Dong, Na and Qin, Sitian and Hu, Xiaolin},
  title = {PartImageNet++ Dataset: Scaling up Part-based Models for Robust Recognition},
  booktitle={European conference on computer vision},
  year = {2024},
  organization={Springer}
}

@article{yang2021ishape,
  title={ishape: A first step towards irregular shape instance segmentation},
  author={Yang, Lei and Wei, Yan Zi and He, Yisheng and Sun, Wei and Huang, Zhenhang and Huang, Haibin and Fan, Haoqiang},
  journal={arXiv preprint arXiv:2109.15068},
  year={2021}
}

@article{xu2018youtube,
  title={Youtube-vos: A large-scale video object segmentation benchmark},
  author={Xu, Ning and Yang, Linjie and Fan, Yuchen and Yue, Dingcheng and Liang, Yuchen and Yang, Jianchao and Huang, Thomas},
  journal={arXiv preprint arXiv:1809.03327},
  year={2018}
}

@inproceedings{silberman2012indoor,
  title={Indoor segmentation and support inference from rgbd images},
  author={Silb, Nathan and Hoiem, Derek and Kohli, Pushmeet and Fergus, Rob},
  booktitle={European conference on computer vision},
  pages={746--760},
  year={2012},
  organization={Springer}
}

@article{decarvalho2022panoptic,
  title={Panoptic Segmentation Meets Remote Sensing},
  author={de Carvalho, Osmar Luiz Ferreira and de Carvalho Junior, Osmar Abilio and e Silva, Cristiano Rosa and de Albuquerque, Anesmar Olino and Santana, Nickolas Castro and Borges, Dibio Leandro and Gomes, Roberto Arnaldo Trancoso and Guimaraes, Renato Fontes},
  journal={Remote Sensing},
  volume={14},
  number={4},
  pages={965},
  year={2022},
  publisher={MDPI},
  doi={10.3390/rs14040965}
}

@InProceedings{zendel2022Unifying,
    author    = {Zendel, Oliver and Sch\"orghuber, Matthias and Rainer, Bernhard and Murschitz, Markus and Beleznai, Csaba},
    title     = {Unifying Panoptic Segmentation for Autonomous Driving},
    booktitle = {Proceedings of the IEEE/CVF Conference on Computer Vision and Pattern Recognition (CVPR)},
    month     = {June},
    year      = {2022},
    pages     = {21351-21360}
}

@article{garnot2021panoptic,
  title={Panoptic Segmentation of Satellite Image Time Series
        with Convolutional Temporal Attention Networks},
  author={Garnot, Vivien and Landrieu, Loic },
  journal={ICCV},
  year={2021}
}

@inproceedings{roberts2021Photorealistic,
    author    = {Mike Roberts AND Jason Ramapuram AND Anurag Ranjan AND Atulit Kumar AND
                 Miguel Angel Bautista AND Nathan Paczan AND Russ Webb AND Joshua M. Susskind},
    title     = {{Hypersim}: {A} Photorealistic Synthetic Dataset for Holistic Indoor Scene Understanding},
    booktitle = {International Conference on Computer Vision (ICCV) 2021},
    year      = {2021}
}

@inproceedings{waqas2019isaid,
title={iSAID: A Large-scale Dataset for Instance Segmentation in Aerial Images},
author={Waqas Zamir, Syed and Arora, Aditya and Gupta, Akshita and Khan, Salman and Sun, Guolei and Shahbaz Khan, Fahad and Zhu, Fan and Shao, Ling and Xia, Gui-Song and Bai, Xiang},
booktitle={Proceedings of the IEEE Conference on Computer Vision and Pattern Recognition Workshops},
pages={28--37},
year={2019}
}

@article{weyler2024pami,
  author = {Jan Weyler and Federico Magistri and Elias Marks and Yue Linn Chong and Matteo Sodano 
    and Gianmarco Roggiolani and Nived Chebrolu and Cyrill Stachniss and Jens Behley},
  title = {{PhenoBench --- A Large Dataset and Benchmarks for Semantic Image Interpretation
    in the Agricultural Domain}},
  journal = {IEEE Trans. on Pattern Analysis and Machine Intelligence (T-PAMI)},
  volume = {46},
  number = {12},
  pages = {9583--9594},
  year = {2024}
}

@inproceedings{dai2017scannet,
    title={ScanNet: Richly-annotated 3D Reconstructions of Indoor Scenes},
    author={Dai, Angela and Chang, Angel X. and Savva, Manolis and Halber, Maciej and Funkhouser, Thomas and Nie{\ss}ner, Matthias},
    booktitle = {Proc. Computer Vision and Pattern Recognition (CVPR), IEEE},
    year = {2017}
}

@article{meletis2020cityscapes,
  title={Cityscapes-panoptic-parts and pascal-panoptic-parts datasets for scene understanding},
  author={Meletis, Panagiotis and Wen, Xiaoxiao and Lu, Chenyang and De Geus, Daan and Dubbelman, Gijs},
  journal={arXiv preprint arXiv:2004.07944},
  year={2020}
}

@inproceedings{bolya2024window,
  title={Window attention is bugged: How not to interpolate position embeddings},
  author={Bolya, Daniel and Ryali, Chaitanya and Hoffman, Judy and Feichtenhofer, Christoph},
  booktitle={International Conference on Learning Representations},
  volume={2024},
  pages={5304--5325},
  year={2024}
}

@inproceedings{heo2024rotary,
  title={Rotary position embedding for vision transformer},
  author={Heo, Byeongho and Park, Song and Han, Dongyoon and Yun, Sangdoo},
  booktitle={European Conference on Computer Vision},
  pages={289--305},
  year={2024},
  organization={Springer}
}

@article{su2104enhanced,
  title={Enhanced transformer with rotary position embedding},
  author={Su, J and Lu, Y and Pan, S and Murtadha, A and Wen, B and Roformer, YL},
  journal={arXiv preprint arXiv:2104.09864},
  year={2021}
}

@inproceedings{jia2020fashionpedia,
  title={Fashionpedia: Ontology, segmentation, and an attribute localization dataset},
  author={Jia, Menglin and Shi, Mengyun and Sirotenko, Mikhail and Cui, Yin and Cardie, Claire and Hariharan, Bharath and Adam, Hartwig and Belongie, Serge},
  booktitle={European conference on computer vision},
  pages={316--332},
  year={2020},
  organization={Springer}
}

@article{hong2020trashcan,
  title={Trashcan: A semantically-segmented dataset towards visual detection of marine debris},
  author={Hong, Jungseok and Fulton, Michael and Sattar, Junaed},
  journal={arXiv preprint arXiv:2007.08097},
  year={2020}
}

@misc{yang2022focal,
  title={Focal Modulation Networks},
  author={Jianwei Yang and Chunyuan Li and Xiyang Dai and Jianfeng Gao},
  journal={Advances in Neural Information Processing Systems (NeurIPS)},
  year={2022}
}

@InProceedings{Ranzinger2024radio,
    author    = {Ranzinger, Mike and Heinrich, Greg and Kautz, Jan and Molchanov, Pavlo},
    title     = {AM-RADIO: Agglomerative Vision Foundation Model Reduce All Domains Into One},
    booktitle = {Proceedings of the IEEE/CVF Conference on Computer Vision and Pattern Recognition (CVPR)},
    month     = {June},
    year      = {2024},
    pages     = {12490-12500}
}

@article{ravi2024sam2,
  title={SAM 2: Segment Anything in Images and Videos},
  author={Ravi, Nikhila and Gabeur, Valentin and Hu, Yuan-Ting and Hu, Ronghang and Ryali, Chaitanya and Ma, Tengyu and Khedr, Haitham and R{\"a}dle, Roman and Rolland, Chloe and Gustafson, Laura and Mintun, Eric and Pan, Junting and Alwala, Kalyan Vasudev and Carion, Nicolas and Wu, Chao-Yuan and Girshick, Ross and Doll{\'a}r, Piotr and Feichtenhofer, Christoph},
  journal={arXiv preprint arXiv:2408.00714},
  url={https://arxiv.org/abs/2408.00714},
  year={2024}
}

@article{liang2026wevisone,
  title={WeVisOne-VL: Unified Visual Perception from Objects to Pixels},
  author={Liang, Tianming and Pan, Junyi and Li, Yuxi and Fu, Canmiao and Huang, Zhipeng and Sun, Chong and Li, Chen and Lyu, Jing},
  journal={arXiv preprint arXiv:2610.31329},
  year={2026}
}

@article{butoi2023universeg,
  title={UniverSeg: Universal Medical Image Segmentation},
  author={Victor Ion Butoi and Jose Javier Gonzalez Ortiz and Tianyu Ma and Mert R. Sabuncu and John Guttag and Adrian V. Dalca},
  journal={International Conference on Computer Vision},
  year={2023}
}

@inproceedings{kirillov2019panoptic,
  title={Panoptic segmentation},
  author={Kirillov, Alexander and He, Kaiming and Girshick, Ross and Rother, Carsten and Doll{\'a}r, Piotr},
  booktitle={2019 IEEE/CVF conference on computer vision and pattern recognition (CVPR)},
  pages={9396--9405},
  year={2019},
  organization={Ieee}
}

@inproceedings{fang2019modularized,
  title={Modularized textual grounding for counterfactual resilience},
  author={Fang, Zhiyuan and Kong, Shu and Fowlkes, Charless and Yang, Yezhou},
  booktitle={IEEE/CVF Conference on Computer Vision and Pattern Recognition (CVPR)},
  year={2019}
}

@inproceedings{cheng2021boundary,
  title={Boundary IoU: Improving object-centric image segmentation evaluation},
  author={Cheng, Bowen and Girshick, Ross and Doll{\'a}r, Piotr and Berg, Alexander C and Kirillov, Alexander},
  booktitle={2021 IEEE/CVF Conference on Computer Vision and Pattern Recognition (CVPR)},
  pages={15329--15337},
  year={2021},
  organization={IEEE}
}

@inproceedings{sun2024alpha,
  title={Alpha-clip: A clip model focusing on wherever you want},
  author={Sun, Zeyi and Fang, Ye and Wu, Tong and Zhang, Pan and Zang, Yuhang and Kong, Shu and Xiong, Yuanjun and Lin, Dahua and Wang, Jiaqi},
  booktitle={IEEE/CVF Conference on Computer Vision and Pattern Recognition (CVPR)},
  year={2024}
}

@article{he2024sprout,
  title={SPROUT: A User-friendly, Scalable Toolkit for Multi-class Segmentation of Volumetric Images},
  author={He, Yichen and Camaiti, Marco and Roberts, Lucy E and Mulqueeney, James M and Ioannou, Eleftherios and Didziokas, Marius and Goswami, Anjali},
  journal={bioRxiv},
  pages={2024--11},
  year={2024},
  publisher={Cold Spring Harbor Laboratory}
}
\bibliographystyle{ieee_fullname}

\appendix
\clearpage

{
\begin{center}
\Large
\textbf{Segment Anything In-Context via Visual Demonstrations}\\
(Appendix)
\end{center}
}

\renewcommand{\thesection}{\Alph{section}}
\renewcommand{\theHsection}{\Alph{section}}
\setcounter{section}{0}

This document supplements our main paper with more details. It is organized as follows: 

\begin{itemize}[
topsep=0.1pt, 
partopsep=0.1pt, 
itemsep=0.1pt,
leftmargin=0.25in,
label={\textbullet}
]

\item {\bf Section \ref{sec_suppl:datasets}} details all the datasets used for training and evaluation.

\item {\bf Section \ref{sec_suppl:arch}} presents model and training details.

\item {\bf Section \ref{sec_suppl:qualitative}} shows more qualitative results.

\item {\bf Section \ref{sec_suppl:code}}  provides code and demos to reproduce our results.

\item {\bf Section \ref{sec_suppl:discussions}} discusses limitations and future work.

\end{itemize}

\section{Datasets}
\label{sec_suppl:datasets}

\begin{table}[bh]
\centering
\vspace{-2mm}
\caption{\small
    \textbf{Training and evaluation datasets}.
    Compared with the 12 training datasets featuring natural scenes, the 24 evaluation datasets are out-of-domain (OOD) that include scientific domains.
    We intentionally incorporate this domain shift in the training-evaluation protocol to rigorously examine models' OOD generalization.
}
\vspace{-3mm}
\scalebox{0.81}{
    \small
    \renewcommand{\arraystretch}{1.1}
    \setlength{\tabcolsep}{1.8pt}
    \begin{tabular}{llrrr}
        \toprule
        dataset \& link & description & mask type & \#images & \#masks\\
       
        \midrule
        \rowcolor{lightlightgrey} \multicolumn{5}{l}{\textit{Training}} \\
        \href{https://groups.csail.mit.edu/vision/datasets/ADE20K/}{ADE20K}~\cite{zhou2017scene} & Natural scenes; object and stuff annotations & Panoptic & 20.2k & 271.2k\\
        \href{https://cocodataset.org/}{COCO}~\cite{lin2014microsoft} & Everyday scenes; thing and stuff annotations & Panoptic & 118.3k & 1,330.0k\\
        \href{https://xdeng7.github.io/coconut.github.io/}{COCONut}~\cite{deng2024coconut} & high-quality, densely annotated COCO extension & Panoptic & 488.3k & 5,836.0k\\
        \href{https://github.com/pmeletis/panoptic_parts}{PASCAL Panoptic}~\cite{meletis2020cityscapes} & Natural scenes; object- and part-level annotations & Panoptic & 10.1k & 69.1k\\
        \href{https://www.mapillary.com/dataset/vistas}{Mapillary Vistas}~\cite{neuhold2017mapillary} & Street-level scenes from geographically diverse locations & Panoptic & 18.0k & 1234.5k\\
        \href{https://www.cityscapes-dataset.com/}{Cityscapes}~\cite{Cordts2016Cityscapes} & Urban street scenes captured from a driving perspective & Panoptic & 3.0k & 79.7k\\
        \href{https://github.com/VIPSeg-Dataset/VIPSeg-Dataset}{VIPSeg}~\cite{miao2022large} & In-the-wild video frames; video panoptic annotations & Panoptic & 75.0k & 828.7k\\
        \href{https://jrdb.erc.monash.edu/dataset}{JRDB-PanoTrack}~\cite{le2024jrdbpanotrack} & Crowded robot-view scenes; panoptic tracking annotations & Panoptic & 18.6k & 397.7k\\
        \href{https://lojzezust.github.io/lars-dataset/}{LaRS}~\cite{Zust2023LaRS} & Maritime scenes; water, sky, and obstacle annotations & Panoptic & 2.6k & 17.0k\\
        \href{https://drive.google.com/drive/folders/1OLBd23ufm6CU8CZmLEYMdF-x2b8mRgxV}{CIHP}~\cite{gong2018instance} & Natural images; instance-aware human-part annotations & Panoptic & 33.3k & 768.4k\\
        \href{https://github.com/tacju/partimagenet}{PartImageNet}~\cite{he2021partimagenet} & ImageNet objects; part-level annotations & Panoptic & 24.1k & 111.8k\\
        \href{https://huggingface.co/datasets/lixiao20/PartImageNetPP}{PartImageNetPP}~\cite{li2024pinpp} & large-scale expansion of PartImageNet & Panoptic & 100.0k & 406.4k\\
        
        \midrule
        \rowcolor{lightlightgrey} \multicolumn{5}{l}{\textit{Evaluation}} \\
        \href{https://download.visinf.tu-darmstadt.de/data/from_games/index.html}{GTA5, GTA}~\cite{richter2016playing} & Synthetic urban driving scenes & Semantic & 6,347 & 80,765\\
        \href{https://faculty.runi.ac.il/arik/site/artseg/Dram-Dataset.html}{DRAM}~\cite{cohen2022semantic} & Figurative paintings from four art movements & Semantic & 144 & 219\\
        \href{https://www.vision.caltech.edu/datasets/cub_200_2011/}{CUB-200-2011, CUB}~\cite{wah2011caltech} & Fine-grained bird images & Semantic & 5,793 & 5,793\\
        \href{http://hcp.sysu.edu.cn/lip}{LIP}~\cite{gong2017look} & Natural images with human-part annotations & Semantic & 10,000 & 72,335\\
        \href{https://youtube-vos.org/}{YouTube-VOS, YouT}~\cite{xu2018youtube} & YouTube video frames with object masks & Semantic & 18,762 & 33,145\\
        \href{https://cs.nyu.edu/~fergus/datasets/nyu_depth_v2.html}{NYU Depth v2, NYU}~\cite{silberman2012indoor} & RGB-D indoor scenes & Semantic & 290 & 2,964\\
        \href{https://ishape.github.io/}{iShape}~\cite{yang2021ishape} & Real and synthetic images of irregularly shaped objects & Semantic & 2,685 & 35,606\\
        
        \href{https://zenodo.org/records/14935094}{PlantSeg, Plant}~\cite{wei2409plantseg} & In-the-wild plant-disease images with lesion masks & Semantic & 1,247 & 1,247\\
        \href{https://calvin-vision.net/datasets/tigdog/}{TigDog, TigD}~\cite{mu2020learning} & Articulated tiger and horse images with part annotations & Semantic & 400 & 3,413\\
        \href{https://www.antscan.info/}{Ant}~\cite{katzke2026high} & Micro-CT-derived views of ant anatomy & Semantic & 96 & 479\\
        \href{https://github.com/junjue-wang/loveda}{LoveDA, LDA}~\cite{Wang2021LoveDA} & High-resolution urban and rural remote-sensing images & Semantic & 1,669 & 7,152\\
        \href{http://db.jsrt.or.jp/eng.php}{JSRT}~\cite{shiraishi2000development} & Chest radiographs with anatomical masks & Semantic & 240 & 720\\
        \href{https://github.com/ari-dasci/OD-MetalDAM}{MetalDAM, ODM}~\cite{metaldam2024} & Metallographic images of additively manufactured materials & Semantic & 36 & 92\\
        \href{https://data.galaxyzoo.org}{Galaxy, Gala}~\cite{willett2013galaxy} & Galaxy Zoo images with author-generated galaxy masks & Semantic & 201 & 226\\
        \href{https://github.com/sartorius-research/LIVECell/tree/main}{LIVECell, Cell}~\cite{edlund2021livecell} & Phase-contrast microscopy images of live cells & Semantic & 569 & 569\\

        \midrule
        \href{https://www.phenobench.org/dataset.html}{PhenoBench}~\cite{weyler2024pami} & UAV field images with crop, weed, and leaf annotations & Instance & 772 & 8,568\\
        \href{https://fashionpedia.github.io/home/Fashionpedia_download.html}{Fashionpedia}~\cite{jia2020fashionpedia} & Fashion images with apparel-instance and part annotations & Instance & 1,158 & 8,781\\
        \href{https://captain-whu.github.io/iSAID/dataset.html}{iSAID}~\cite{waqas2019isaid} & High-resolution aerial images with object-instance masks & Instance & 349 & 99,499\\
        \href{https://kaldir.vc.in.tum.de/scannet_benchmark/documentation}{ScanNet}~\cite{dai2017scannet} & RGB-D indoor scenes reconstructed from video & Instance & 520 & 9,764\\
        \href{https://conservancy.umn.edu/items/6dd6a960-c44a-4510-a679-efb8c82ebfb7}{TrashCan}~\cite{hong2020trashcan} & Underwater images with marine-debris annotations & Instance & 1,147 & 2,588\\
        
        \midrule
        \href{https://wilddash.cc/download}{WildDash 2}~\cite{zendel2022Unifying} & Diverse and adverse road-driving scenes & Panoptic & 638 & 21,353\\
        \href{https://huggingface.co/datasets/osmarluiz/BSB-Aerial-Dataset}{BSB}~\cite{decarvalho2022panoptic} & High-resolution remote-sensing scenes with building instances & Panoptic & 200 & 9458\\
        \href{https://github.com/VSainteuf/pastis-benchmark}{PASTIS}~\cite{garnot2021panoptic} & Satellite image time series of agricultural parcels & Panoptic & 487 & 16,940\\
        \href{https://github.com/apple/ml-hypersim/tree/main}{Hypersim}~\cite{roberts2021Photorealistic} & Photorealistic synthetic indoor scenes & Panoptic & 178 & 2,261\\
        \bottomrule
    \end{tabular}
}
\vspace{-1mm}
\label{tab:appendix-dataset}
\end{table}

\begin{figure}[t] 
    \centering
    \includegraphics[width=0.99\textwidth]{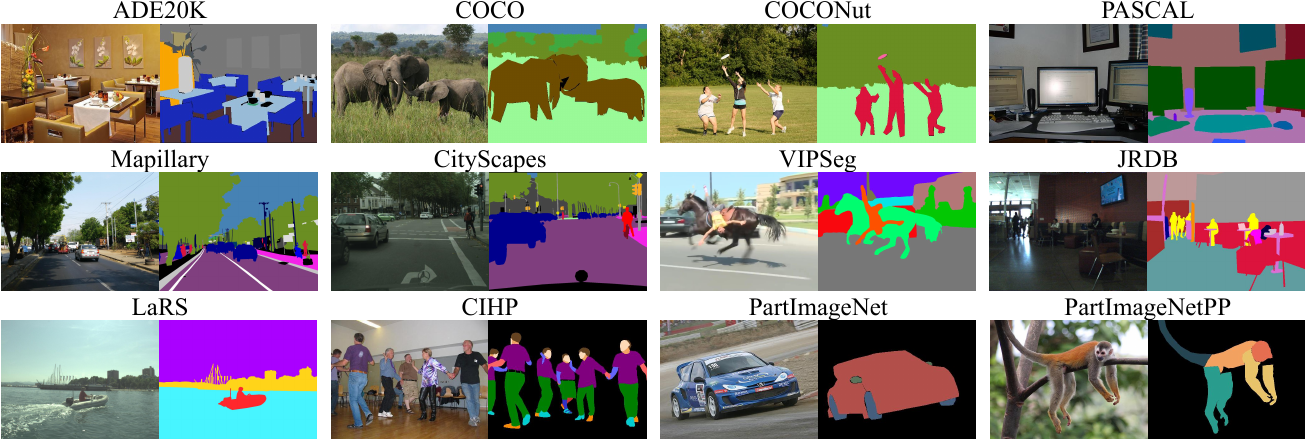}
    \vspace{-2.5mm}
    \caption{\small
    \textbf{Random examples from the 12 training datasets},
    including images and their color-coded semantic segmentation masks.
    }
    \label{fig:appendix-train-sample}
\end{figure}

\textbf{Training datasets.} 
We train \ourmodel on 12 publicly-available datasets originally curated for panoptic segmentation.
These datasets capture images of natural scenes.
Collectively, they provide 0.91M images and 11.46M annotated masks.
Table~\ref{tab:appendix-dataset} lists the datasets, their domains, annotation types, and numbers of images and masks. 
Fig.~\ref{fig:appendix-train-sample} displays random training examples from the 12 training datasets.
Notably, SAM3~\cite{carion2025sam} is trained on much more data, including 53.9M images and 1.52B mask-noun phrase pairs.

\begin{figure}[t] 
\centering
    \includegraphics[width=0.99\textwidth]{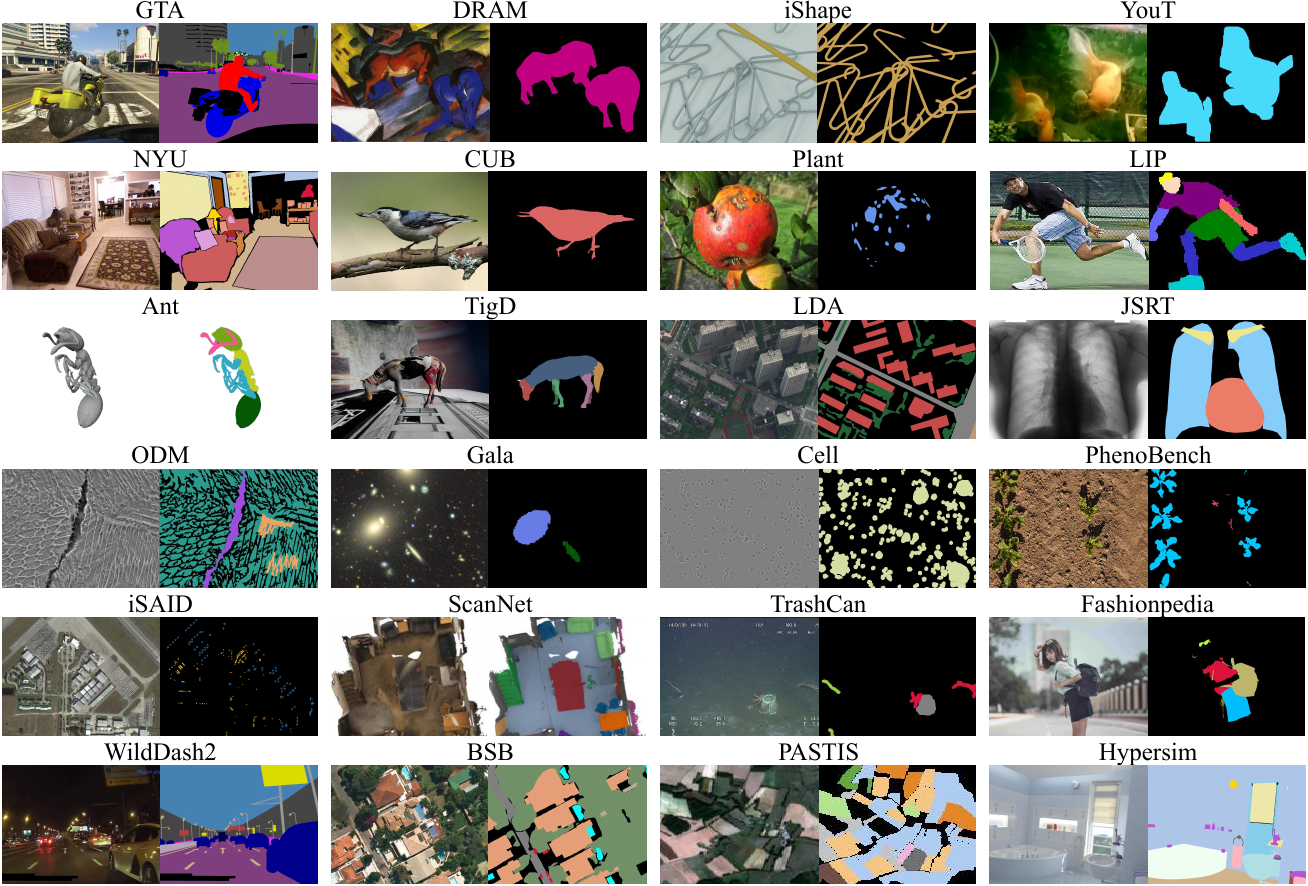}
    \vspace{-3.5mm}
    \caption{\small
    \textbf{Random examples from the 24 evaluation datasets},
    featuring images from diverse domains.
    }
    \label{fig:appendix-eval-sample}
\end{figure}

\textbf{Evaluation datasets.}
We use  24 datasets to benchmark different methods.
These datasets contain diverse domains, such as biology, medical imaging, remote sensing, astronomy, agriculture, manufacturing, and synthetic environments, pathology, and art.
They have different levels of segment granularity, either object-centric or part-focused.
Table~\ref{tab:appendix-dataset} lists their dataset statistics.
Fig.~\ref{fig:appendix-eval-sample} shows random examples from these datasets.
For efficient benchmarking, we use subsets of the ScanNet and Hypersim validation sets.
The Galaxy dataset contains images sampled the Galaxy Zoo repository.

\section{More Details of Model Architecture and Training}
\label{sec_suppl:arch}

\textbf{Image encoder.}
Following \cite{li2022exploring}, we employ a pre-trained ViT-L~\cite{dosovitskiy2020image, simeoni2025dinov3, yang2026pursuit, bolya2026perception} as the image encoder, 
incorporating 16$\times$16 windowed attention and 4 evenly spaced global attention blocks to process input images at 512$\times$512 resolution. 
We discard the \texttt{[class]} token and adopt a 16$\times$16 patch size, yielding 1,024 patch tokens per image.
For $\text{PE}_{\text{spatial}}$,
the pretrained patch-embedding filters are interpolated from 14$\times$14$\times$3 to 16$\times$16$\times$3.
We retain the positional-encoding scheme associated with each pretrained encoder, 
either windowed absolute positional embeddings~\cite{bolya2024window} or RoPE~\cite{su2104enhanced, heo2024rotary}.
The positional encoding is applied independently to patch tokens within each image and shared across the query and prompts;
no positional index is constructed across images. The encoder can therefore process a variable number of prompt images at inference time.

\textbf{Mask decoder.}
The mask decoder follows a query-based segmentation architecture~\cite{cheng2021per, cheng2022masked}. 
It receives single-scale, 16$\times$ downsampled features and a set of learnable mask queries, and outputs paired masks and class scores.
As illustrated in Fig.~\ref{fig:architecture}(a), each decoder block sequentially applies: (1) masked cross attention from mask queries (as queries) to the query image feature (taken from the final layer of the image encoder), (2) self-attention among the mask queries, and (3) a pointwise MLP updates each mask query. In step (1), we adopt the masked attention mechanism proposed in Mask2Former~\cite{cheng2022masked}.
Prompt features are used \emph{only} for prototype-based mask classification and do not enter the mask prediction path. 
The decoder contains 4 Transformer blocks, with embedding dimension 256,
8 attention heads, and 200 mask queries. 
To preserve spatial information during decoding, positional embeddings are re-injected at every decoder block: 
learnable embeddings are added to the mask queries~\cite{cheng2022masked}, 
while fixed sinusoidal embeddings~\cite{he2022masked} are added to the query image features only when these features serve as keys (rather than values) in the masked cross-attention.

\begin{wrapfigure}{r}{0.42\textwidth}
{
\centering
\vspace{-0mm}
\captionof{table}{\small 
\textbf{Architectural specifications of all compared methods.}
$^\dagger$: Short side resized to 800 px; long side capped at 1,333 px.
}
\label{tab:appendix-methods}
\vspace{-3.3mm}
\scalebox{0.8}{
    \renewcommand{\arraystretch}{1.1}
    \setlength{\tabcolsep}{1.5pt}
    \begin{tabular}{l *{3}{r}}
        \toprule
        Model & Image encoder & Params (M) & Image size\\  
               
        \midrule

        X-Dec \tiny\cite{zou2023generalized} & Focal-L & 341.6 & 512$^2$\\
        
        O-SeeD \tiny\cite{zhang2023simple} & Swin-T & 116.2 & 512$^2$\\
        
        APE-D \tiny\cite{shen2024aligning} & ViT-L & 1,074.7 & 800$^\dagger$\\
        
        G-SAM \tiny\cite{ren2024grounded} & Swin-B & 396.8 & 1,024$^2$\\

        O-SAM \tiny\cite{xiao2025openworldsam} & ViT-H & 902.9 & 1,024$^2$\\

        R-SAM \tiny\cite{alama2026radseg} & C-RADIOv3-B & 493.1 & 896$^2$\\

        SAM3 \tiny\cite{carion2025sam} & ViT-H & 840.5 & 1,008$^2$\\

        Painter \tiny\cite{wang2023images} & ViT-L & 310.7 & 448$^2$\\
        
        DINOv \tiny\cite{li2024visual} & Swin-L & 224.6 & 640$^2$\\

        INSID3 \tiny\cite{cuttano2026insid3} & ViT-L & 303.2 & 1,024$^2$\\
        
        \rowcolor{col1} 
        \ourmodel (ours) & ViT-L & 310.7 & 512$^2$\\
        \bottomrule
    \end{tabular}
}
\vspace{-2mm}
}
\end{wrapfigure}
\textbf{Specifics of different models.}
Table~\ref{tab:appendix-methods} summarizes the image encoders, parameter counts, and default input input resolutions of all compared methods in our experiments.
Most methods use large backbones:
ViT-L~\cite{dosovitskiy2020image}, Swin-L~\cite{liu2021swin}, or Focal-L~\cite{yang2022focal};
O-SAM and SAM3 further scale up to ViT-H. 
O-SeeD and G-SAM use the largest publicly available variants reported by their authors; and R-SAM uses C-RADIOv3-B~\cite{Ranzinger2024radio}, which is reported to outperform its Large counterpart (see Table~3 of~\cite{alama2026radseg}). 
All qualitative and quantitative comparisons in this paper use each method at its default input resolution.

\textbf{Prototype extraction.}
Suppose that a single prompt image is given.
The prototype for class-$i$, denoted as $\mathbf{p}_i \in \mathbb{R}^{1\times d}$, 
is computed by aggregating the image's feature map $\mathbf{E}$ within its corresponding mask $\mathcal{M}_i$:
$\mathbf{p}_i = \frac{\mathcal{M}_i \odot \mathbf{E}}{|\mathcal{M}_i|} \in \mathbb{R}^{1\times d}$.
Here, $\mathcal{M}_i$ is subsmapled appropriately to match the spatial resolution of $\mathbf{E} \in \mathbb{R}^{h\times w \times d}$.
Finally, each extracted prototype $\mathbf{p}_i$ is $\ell_2$-normalized.
All the $K$ prototypes, defined by the visual demos,
are written as matrix $\mathbf{P}\in \mathbb{R}^{K\times d}$.

{\bf Bias and temperature initialization.} For a mask query and a class prototype, let $s$ denote their cosine similarity. The scaled, calibrated logit is $\tilde s = \frac{s}{\tau} + \beta$. 
At the beginning of training, $s$ is close to zero, which pushes the sigmoid probability to around 0.5. 
As a result, the BCE loss is dominated by many negative samples, which include mask queries that are either left unmatched or assigned to the \textit{ignota} category via the Hungarian algorithm~\cite{carion2020end}.
To mitigate this, we initialize the bias term $\beta$ to -4.59, reducing the initial sigmoid probability to 0.01. 
As for the temperature $\tau$, following standard contrastive learning practice~\cite{wu2018unsupervised, he2020momentum}, we initialize it to 0.07 and clip $1/ \tau$ to [10, 20] during training.

\textbf{Loss.}
We supervise mask prediction using established mask loss~\cite{cheng2021per}, which is the combination of focal loss $\mathcal{L}_{\text{focal}}$ and dice loss $\mathcal{L}_{\text{dice}}$.
For mask classification, we adopt a sigmoid-based binary cross-entropy loss $\mathcal{L}_{\text{bce}}$, augmented with focal modulation ($\alpha=0.5, \gamma=1.0$)~\cite{lin2017focal} to more effectively handle the imbalance between positive and negative mask queries.

\textbf{Loss weight tuning.}
To determine the loss weights for the loss terms $\mathcal{L}_{\text{dice}}$,
$\mathcal{L}_{\text{focal}}$,
and $\mathcal{L}_{\text{bce}}$,
we first conduct a pilot experiment to train \ourmodel for 100k iterations with all the weights as 1.
Empirically, the mask prediction loss terms $\mathcal{L}_{\text{dice}}$ and
$\mathcal{L}_{\text{focal}}$ converge substantially faster than the classification loss $\mathcal{L}_{\text{bce}}$.
Then, we focus on the weight for $\mathcal{L}_{\text{bce}}$.
We consider the magnitude (i.e., the $L_2$ norm) of gradient produced by the three terms:
with weight 1 for $\mathcal{L}_{\text{dice}}$ and
$\mathcal{L}_{\text{focal}}$,
we increase the weight for $\mathcal{L}_{\text{bce}}$ until 
it produces gradient magnitude (after 100k training iterations) that matches the summed gradient magnitude of $\mathcal{L}_{\text{dice}}$ and
$\mathcal{L}_{\text{focal}}$.
Then, we adjust the weight for $\mathcal{L}_{\text{focal}}$ to match the gradient magnitude of $\mathcal{L}_{\text{dice}}$.
Consequently, 
the final weights for the loss terms $\mathcal{L}_{\text{dice}}$,
$\mathcal{L}_{\text{focal}}$,
and $\mathcal{L}_{\text{bce}}$ are 1, 2, and 20, respectively.
In other word, the final loss $\mathcal{L} = 20 \cdot \mathcal{L}_{\text{bce}} + 2 \cdot \mathcal{L}_{\text{focal}} + \mathcal{L}_{\text{dice}}$.
The loss weights are fixed throughout training. 
Lastly, 
we apply the same loss for deep supervision after each decoder block~\cite{cheng2021per, carion2020end} during training.

{\bf Inference calibration and rejection.} For each predicted mask, we first 
take the maximum value from the calibrated classification logit over the $K$ prompt-defined categories. 
The mask is classified as \textit{ignota} if the max value is less than 0;
otherwise, it is assigned to the class corresponding to this max value.

\begin{figure}[t] 
\centering
    \includegraphics[width=0.99\textwidth]{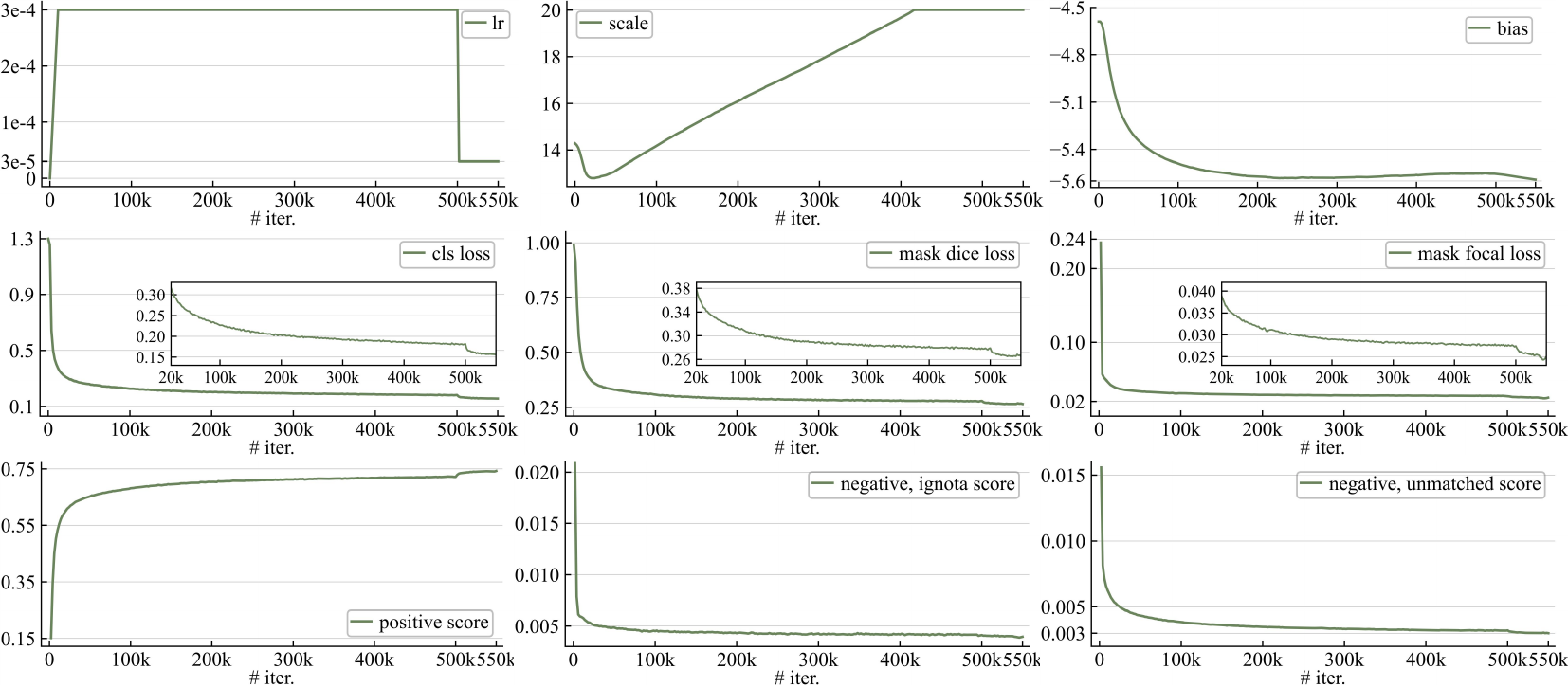}
    \vspace{-2mm}
    \caption{\small
    \textbf{Training curves of key metrics.} Progression of optimization hyperparameters (top row), loss components (middle row), and mean sigmoid confidence scores across positive, \textit{ignota}-matched negative, and unmatched negative mask queries (bottom row) over training iterations.
    }
    \label{fig:appendix-monitor}
\end{figure}

\textbf{Training curves.} Fig.~\ref{fig:appendix-monitor} illustrates the progression of key metrics throughout training.
Specifically, we monitor optimization parameters including the learning rate, alongside the scale ($1/\tau$) and bias term $\beta$ for the classification logits. 
We also plot the individual loss components scaled by their respective weights: the classification loss $\mathcal{L}_{\text{bce}}$, the mask dice loss $\mathcal{L}_{\text{dice}}$, and the mask focal loss $\mathcal{L}_{\text{focal}}$. 
Furthermore, we record the mean sigmoid scores across three distinct subsets of mask queries: positive queries, negative queries matched to the \textit{ignota} category, and unmatched negative queries.

\begin{figure}[h]
\centering
\includegraphics[width=0.99\textwidth]{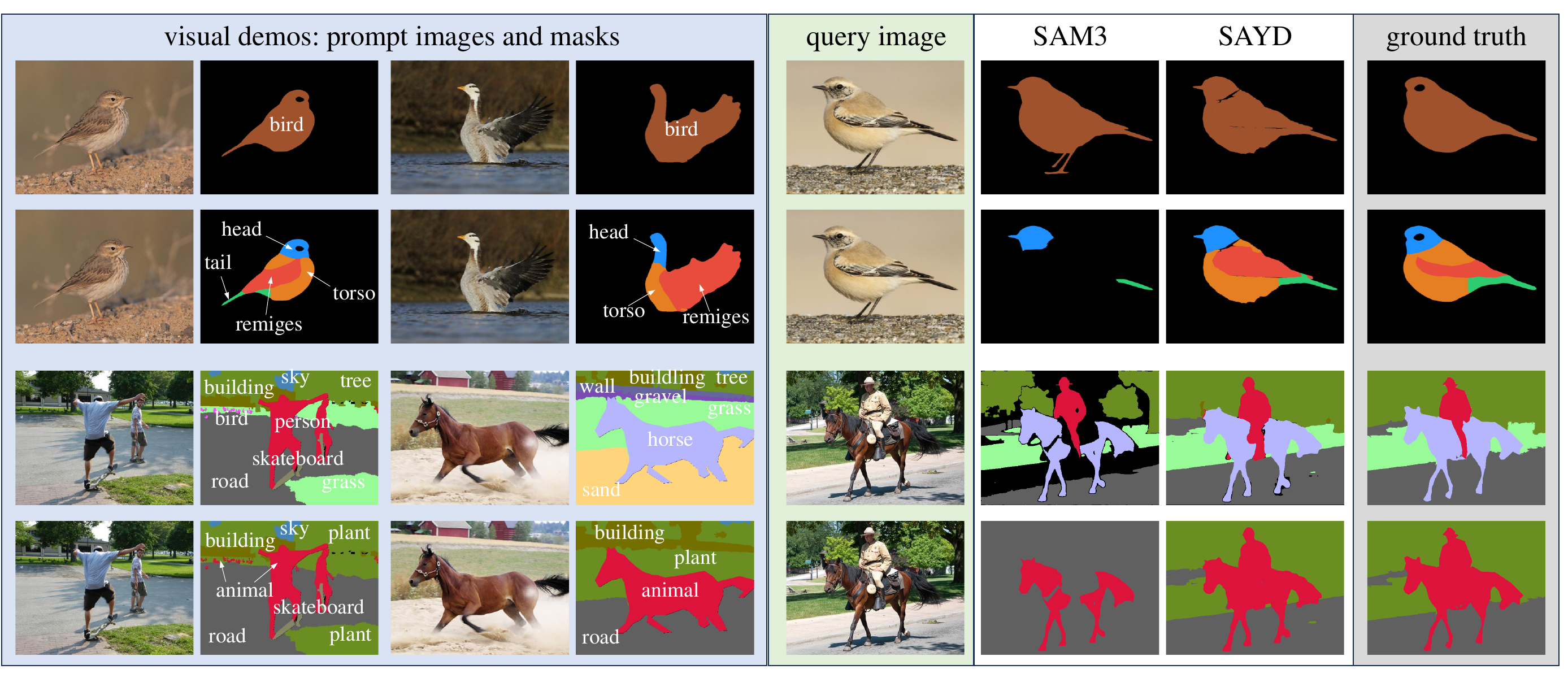}
\vspace{-2.mm}
\caption{\small
\textbf{Qualitative comparison between SAYD and SAM3 at different levels of segmentation granularity.}
Guided by visual demos, our \ourmodel effectively segments the query image at both the object and the part levels. 
In contrast, SAM3, requiring text prompts,
struggles to segment the query images according to customized requirements and shows an inductive bias toward common concepts.
}
\vspace{-0mm}
\label{fig:appendix-taict}
\end{figure}

\section{More Qualitative Results}
\label{sec_suppl:qualitative}

\textbf{Segmentation at different levels of granularity.}
Fig.~\ref{fig:appendix-taict} compares \ourmodel and SAM3 at two different levels of granularity: object-level and part-level, for the same query images.
\ourmodel is conditioned on visual demonstrations, whereas SAM3 is prompted using category names.
The qualitative comparison shows that \ourmodel effectively segment query images by following instructions from the visual demos,
while the SAM3's segmentation results do not consistently match the desired segmentation results. 
Rather, SAM3 shows an inductive bias toward common concepts and fails to segment images according to customized requirements.

\begin{figure}[t]
\centering
\includegraphics[width=0.99\textwidth]{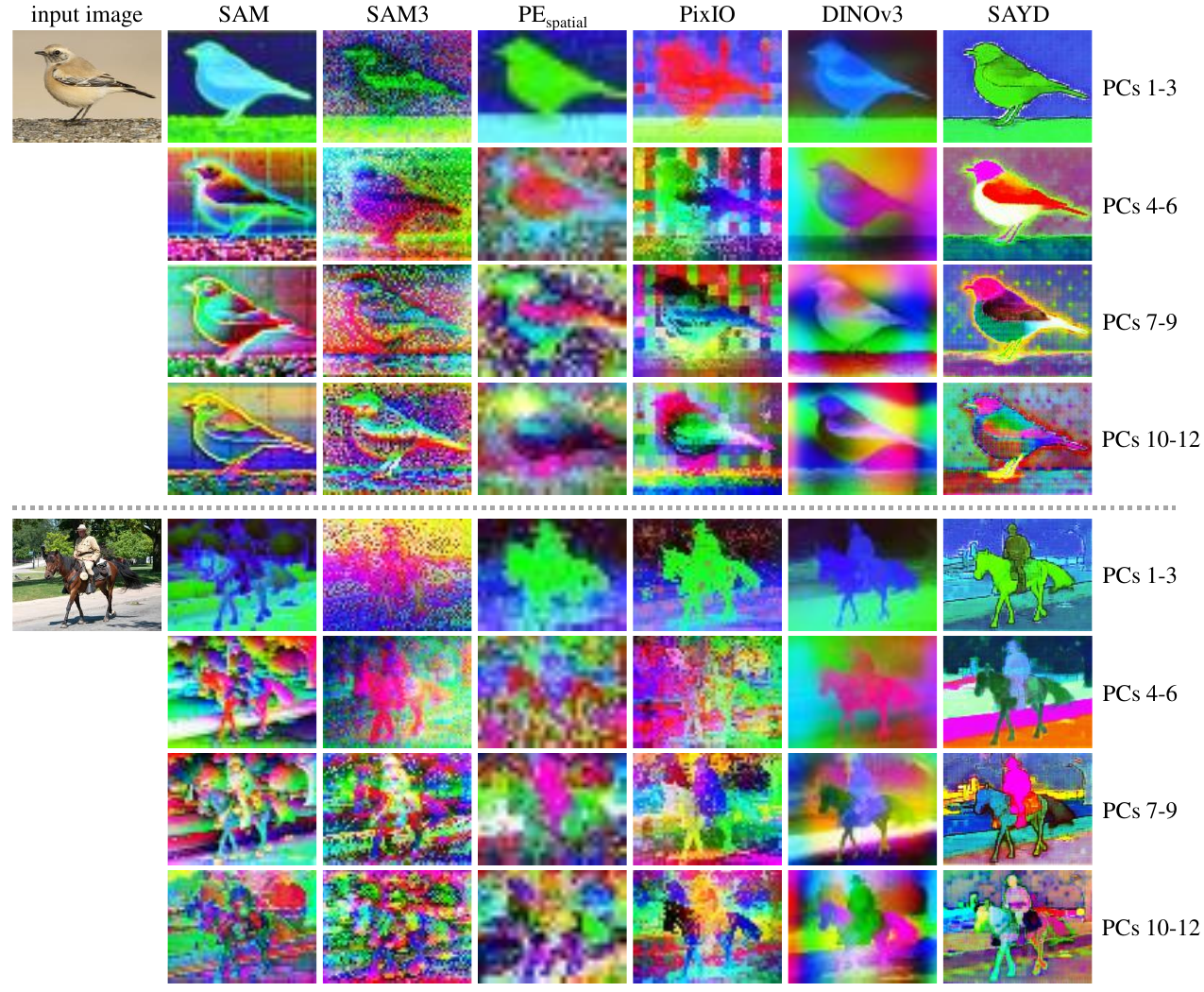}
\vspace{-3.mm}
\caption{\small
{\bf PCA visualization of feature maps computed by different models.}
Rows show different groups of principal components mapped to RGB space.
The feature maps produced by \ourmodel exhibit notably coherent regions
at multiple levels of granularity, from object-level to
part-level separation.
In contrast, the feature maps from the compared models do not imply clear separation of meaningful objects and parts.
}
\vspace{-3mm}
\label{fig:appendix-pca}
\end{figure}

\textbf{Visualization of feature maps via PCA.}
Fig.~\ref{fig:appendix-pca} visualizes principal-component projections of dense feature maps extracted from the final layer of each model's image encoder. SAM3~\cite{carion2025sam} and $\text{PE}_{\text{spatial}}$~\cite{bolya2026perception} are vision–language models, whereas  SAM~\cite{kirillov2023segment}, PixIO~\cite{yang2026pursuit}, DINOv3~\cite{simeoni2025dinov3} and \ourmodel are vision foundation models.
Compared with the other representations, the feature map produced by our \ourmodel  exhibit more visible separations between objects, parts, and background regions.
The visual comparison demonstrates the effecacy of \ourmodel for ICS at different levels of granularity.

\begin{figure}[t]
    \centering
    \vspace{-3.5mm}
        \includegraphics[width=0.99\textwidth]{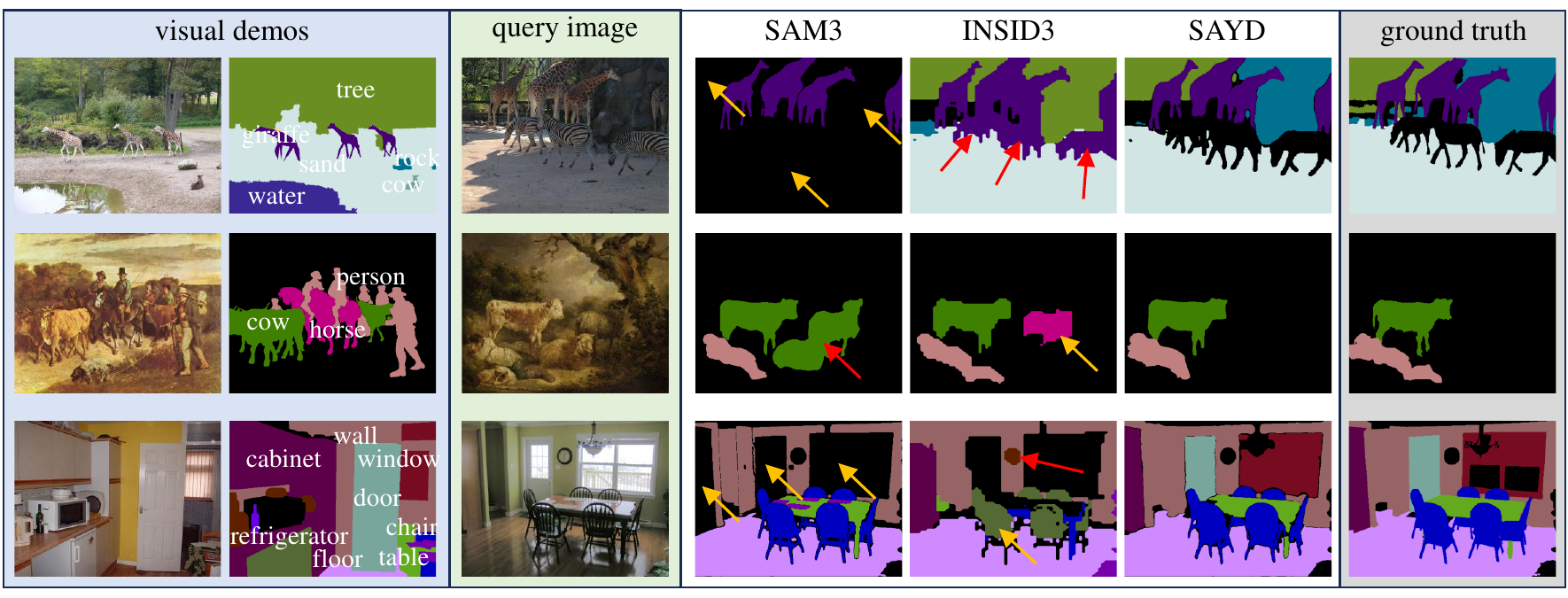}
\vspace{-3.mm}
\caption{\small
\textbf{Visual comparison of results by different 
methods in terms of effectively identifying \textit{ignota}.}
Recall that \textit{ignota} arise when the query image contain novel object classes absent from the visual demos.
In this case, robust models should identify regions of \textit{ignota}, i.e., predicting them using black color along with background regions by our design.
The comparison demonstrates the superior performance of \ourmodel in handling \textit{ignota} over the other two methods.
}
\vspace{-3mm}
\label{fig:appendix-ignota}
\end{figure}

\textbf{Visual comparison of ICS results by different methods in terms of identifying \textit{ignota}.}
Fig.~\ref{fig:appendix-ignota} presents a visual comparison of ICS results by SAM3, INSID3, and our \ourmodel and highlight their ability of identifying \textit{ignota},
i.e., regions of novel concepts absent from visual demos.
In the first row, 
INSID3 fails to detect the \textit{ignota}, 
which are zebras not present in visual demos (pointed by red arrows).
SAM3 does not fire on the \textit{ignota} but fails to identify the known regions such as tree, sand, and water (pointed by orange arrows).
Our \ourmodel effectively identifies them and mark them in black.
In the second row,
SAM3 fires on \textit{ignota} (which belong to sheep not presented in the visual demos; pointed by a red arrow), classifying them as cows.
INSID3 misclassifies them as horse (pointed by an orange arrow).
Our \ourmodel effectively identifies them as \textit{ignota} and mark them in black.
In the third row, 
INSID3 fires on \textit{ignota} which belong to a wall clock (absent from the visual demos; pointed by a red arrow);
SAM3 misclassifies many known classes as background or \textit{ignota} such as window and cabinet.
\ourmodel successfully and accurately handles all these regions.

\begin{figure}[t]
    \centering
        \includegraphics[width=0.99\textwidth]{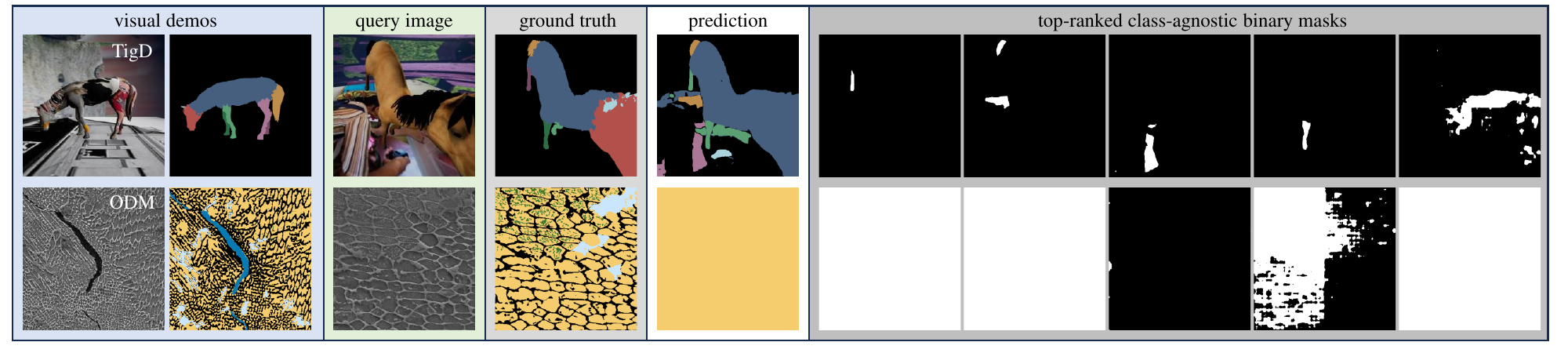}
\vspace{-3.mm}
\caption{\small
\textbf{Failure cases.}
\ourmodel does not output high-quality masks for target regions which feature objects that are rarely seen or of a irregular shape. Consequently, it cannot segment the query images well.
}
\vspace{-3mm}
\label{fig:appendix-failure}
\end{figure}

\textbf{Analysis of failure cases.}
Fig.~\ref{fig:appendix-failure} displays some failure cases of \ourmodel from the TigD and ODM benchmarks.
The first example shows that \ourmodel fails to segment parts of an artistic horse which is captured from an unusual camera view.
The rare observation challenges \ourmodel.
The second example features a metallographic image of additively manufactured materials.
The ground-truth segment masks are highly irregular,
for which \ourmodel fails to generate high-quality masks.
In both cases,
\ourmodel does not output high-quality masks for the target regions.
This implies a limitation of the mask decoder,
which generate binary masks solely based on query images.

\section{Open-Source Code Release}
\label{sec_suppl:code}

{\bf Code.}
We supplement our manuscript with a self-contained code repository (refer to the zip file {\tt Code-SAYD.zip}).
We refer the reader to {\tt README.md}  for instructions how to use the code.
Due to limited space,
we do not upload our trained model (ref. \ourmodel takes $>$200MB physical storage).
We will release both the code and models, under the MIT License, to the public to foster future research.

{\bf Requirement}.
Running our {\tt Python} code requires some common packages, such as Torch, TorchVision, and Timm. Please refer to {\tt Code-SAYD/README.md} for more details.

{\bf Demo.}
We provide three Jupyter Notebook files that demonstrate performance evaluation, prediction visualization on given visual demos,
and PCA visualization of feature maps. 
With these files, the reader can open them and have a better understanding regarding  evaluation and visualization.
We refer the reader to these files: {\tt demo-SAYD-eval.ipynb}, {\tt demo-SAYD-pred.ipynb}, and {\tt demo-SAYD-pca.ipynb} for more details.

\section{Limitations and Future Work}
\label{sec_suppl:discussions}

We discuss several limitations in this work that are worthy of address in future research.
First, while our vision foundation model \ourmodel significantly outperforms existing methods on out-of-domain benchmarks, particularly scientific imagery,
its metrics remain notably lower than on benchmarks featuring common concepts.
Although training \ourmodel on more scientific data could close this gap,
we intentionally maintain a clear domain gap between training and evaluation to rigorously study out-of-domain ICS generalization.
We leave domain-specific adaptation for future work.
Second, 
while our simplified mask decoder helps \ourmodel achieve superior performance over other methods and improve its computational efficiency,
its class-agnostic mask generation nature may struggle to segment irregular target (e.g., the failure case on the MetalDAM ODM benchmark), whereas class-aware mask generation mechanism (e.g., by conditioning on prompts) may mitigate this limitation.
Future work can investigate new decoder designs.
Third,
each visual concept (e.g., object class) is currently represented by a single prototype feature.
Real-world applications may involve complex observations that require a ``bag of features'' to adequately represent a single concept;
developing methodologies to address this is left for future research.
Lastly,
while \ourmodel demonstrates promising segmentation results on images from scientific domains
its practical deployment in these domains to assist domain experts and  accelerate scientific discovery represents a valuable next step.


\end{document}